\documentclass[a4paper,twocolumn,twoside]{article}
\usepackage{amsmath,amssymb}
\usepackage[english]{babel}
\usepackage[small]{caption}
\usepackage{color}
\usepackage{enumitem}
\usepackage[T1]{fontenc}
\usepackage[pdftex]{graphicx}
\usepackage{hyperref}
\usepackage{iggsc}
\usepackage{multirow}
\usepackage{siunitx}

\definecolor{linkcolor}{RGB}{120,0,0}
\newcommand{\TheTitle}[1]{A New Approach to an Old Problem: the Reconstruction{#1}of a Go Game through a Series of Photographs.}
\newcommand{\SssSec}[1]{\vspace{3pt}\noindent\refstepcounter{SssSec}\textbf{\theSssSec\quad#1}}
\newcommand{\ProsCons}[2]{\noindent\textit{\textbf{The pros}: #1}\newline\noindent\textit{\textbf{The cons}: #2}}
\newcommand{\HalfWidth}{.471\columnwidth}
\newcommand{\FullWidth}{\columnwidth}
\newcommand{\EatDot}[1]{}
\newcounter{SssSec}[subsubsection]
\hypersetup{
  pdftitle={\TheTitle{ }},
  pdfauthor={Andrea Carta and Mario Corsolini},
  pdfsubject={PhotoKifu},
  colorlinks=true,
  citecolor=[RGB]{0,80,0},
  linkcolor=[RGB]{120,0,0},
  urlcolor=[RGB]{0,0,160}}

\begin{document}

\author{\normalfont{Andrea Carta\thanks{~Both authors contributed equally.}} \\ \normalfont{\url{http://www.micini.net/}} \and \normalfont{Mario Corsolini{\color{linkcolor}{\textsuperscript{$*$}}}} \\ \normalfont{\url{http://www.oipaz.net/}}}
\title{\TheTitle{\\}}
\date{}
\maketitle

\phantomsection
\addcontentsline{toc}{section}{Abstract}
\begin{abstract}
  Given a series of photographs taken during a Go game, we describe the techniques we successfully employ for pinpointing the grid lines of the Go board and for tracking their small movements between consecutive photographs; then we discuss how to approximate the location and orientation of the observer's point of view, in order to compensate for projection effects. Finally we describe the different criteria that jointly form the algorithm for stones' detection, thus enabling us to automatically reconstruct the whole move sequence.
\end{abstract}

\section{Introduction}

The identification of a Go position by means of a photograph (or a still frame) has been widely dealt with in the last 10 years. In theory, the complete score of a Go game could be reconstructed if enough photos are available: a remarkable feat that could prove very useful given the persistent lack of touch-sensitive ``gobans''.\!\footnote{~Goban is the Japanese term for Go board.} While it is true that under ideal conditions the identification of a position is not a difficult problem (nor it is an easy one), things change a lot if we try to analyse a whole game, often under bad conditions, such as a low point of view, a faint environment light, the presence of shadows/reflections and the continual presence of the players' hands between the camera and the goban. Furthermore, the game reconstruction should be completed in a matter of minutes or even in real time, otherwise the traditional way (manual compilation of a ``kifu''\footnote{~A kifu is a pre-printed grid where the game is recorded.}) would still be preferable to an automated task of a sort.

That's likely the reason why most attempts to this day did not go further the test phase or disappeared completely from the Internet after some time.\!\footnote{~For instance: \textit{AutoGoRecorder}~\cite{CPR11}, \textit{Go Game Recorder}~\cite{Ecl13}, \textit{Go Watcher}~\cite{Gow09}, \textit{GoTracer}~\cite{VV09}, \textit{Image2SGF}~\cite{Bal04}, \textit{kifu}~\cite{Sna09}, \textit{Rocamgo}~\cite{dlCV12}.} The first known program capable of detecting the stones in a single image was likely \textit{CompoGo}~\cite{Boa03}, although it needed optimal conditions; \textit{GoCam}~\cite{Hir05} was likely the first attempt at analysing a video and was capable of detecting the grid, but was never completed; the same happened to \textit{Saikifu}~\cite{Eng07}, that looked promising, but has been given up some years ago. In 2007 Alexander Seewald~\cite{See09} wrote the most interesting theoretical analysis of the problem, claiming a high success rate on single images, but did not develop a program capable of analysing a whole game. In the following years many other studies looked promising, but no software was ever developed despite some good theoretical works: among them only \textit{Webcam~+~Go}~\cite{Seb10} went a bit further, but the author himself admitted it could not analyse a whole game. Eventually something interesting appeared: first \textit{Kifu Snap}~\cite{Cou13}, a not-freeware program (for Android) capable of correctly analysing most single pictures; then \textit{Imago}~\cite{Mus14}, a program that at last was capable of analysing a whole game without making too many errors (its success rate is about 76\%). Unfortunately it takes about 30 seconds per photo, according to the author himself.

Despite a situation not looking very promising, with only one program --- \textit{Imago} --- really capable (albeit slowly) of analysing a whole game, we'll show that it is possible indeed to achieve good results at the speed of a fraction of second per move, even under less-than-ideal conditions. In section~\ref{sec:TrackingGrid} we describe how to locate the grid lines of a goban in the first photo; thereafter we expound how to swiftly follow, in the sequence of the photos, the small displacements of the grid caused by movements of the goban and/or movements of the camera; then, for each photo, we explain how to infer the position and orientation of the camera. In section~\ref{sec:DetectingStones} we give full details of how to make the most of previously collected information and several other criteria to detect the stones put on the goban. In section~\ref{sec:Conclusion} we draw conclusions and outline plausible future developments.

\section{Tracking the grid}
\label{sec:TrackingGrid}

\subsection{Starting location of the grid}
\label{subsec:StartingGrid}

\begin{figure}[!htb]
  \centering
  \setlength{\unitlength}{\FullWidth}
  \begin{picture}(1,0.666667)
    \put(0,0){\includegraphics[width=\FullWidth]{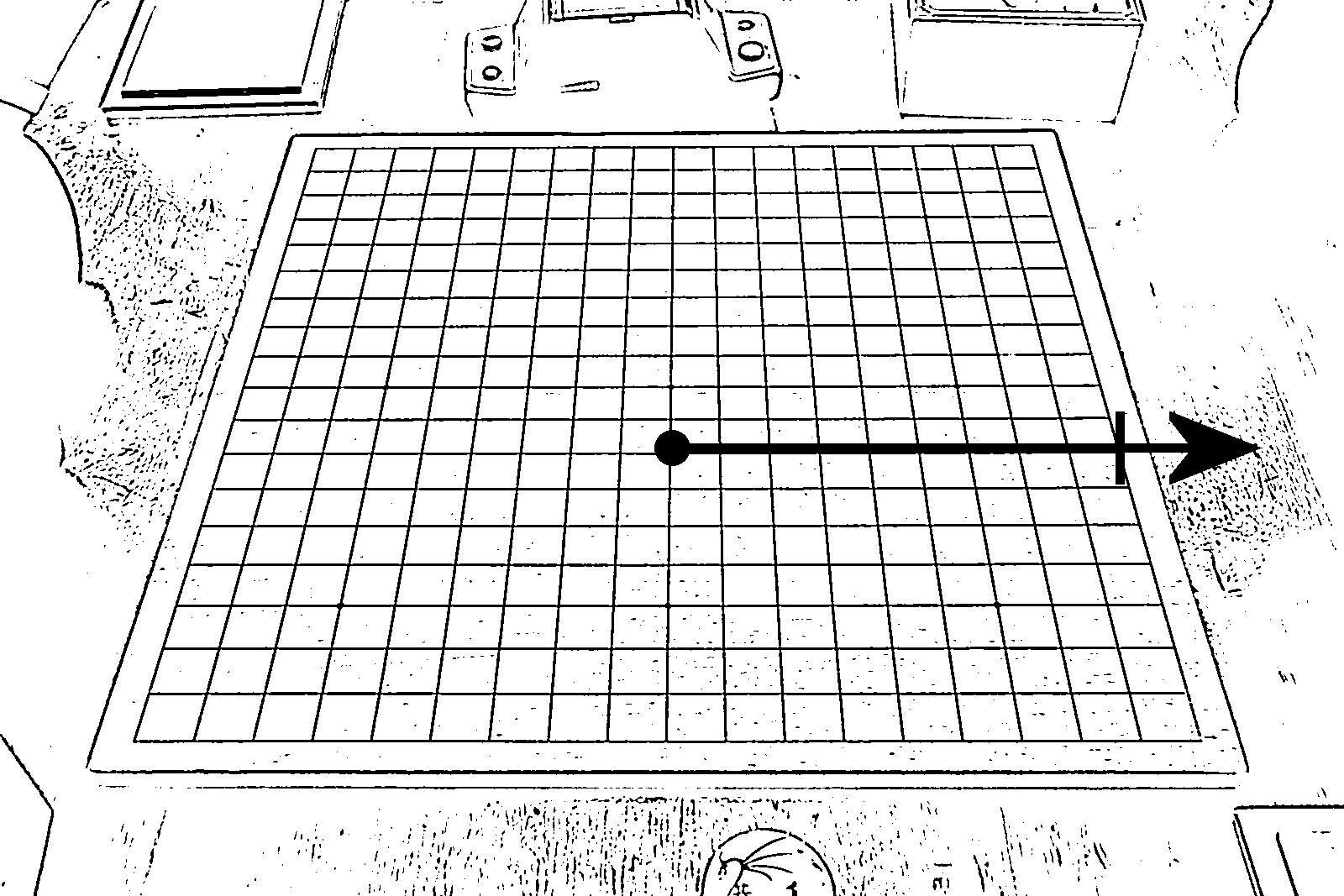}}
    \put(0,0){\line(1,0){1}}
    \put(0,0){\line(0,1){0.666667}}
    \put(1,0){\line(0,1){0.666667}}
    \put(0,0.666667){\line(1,0){1}}
  \end{picture}
  \caption{difference of Gaussians and B/W conversion of a photo of a goban. The dot in the centre is the pole of the polar coordinate system needed in paragraphs \ref{par:HT_MainHough}--\ref{par:HT_Interpolation}; the unit of measure for radial coordinates is marked in the polar axis and it coincides with half of the height of the photo (thus the diagonal is $\sqrt{13}$ units, for full frame digital cameras).}
  \label{fig:HT_DifferenceOfGaussians}
\end{figure}
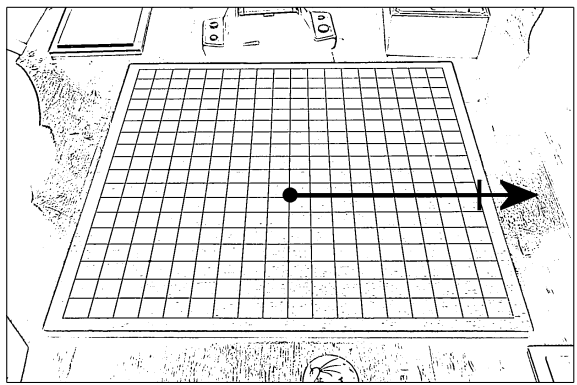

\SssSec{} A ``difference of Gaussians'' filter is applied to the first photo of the series, which is then converted into a B/W image --- one bit per pixel --- in order to highlight all the visible edges, as shown in figure~\ref{fig:HT_DifferenceOfGaussians}.

\begin{figure}[!htb]
  \centering
  \setlength{\unitlength}{\FullWidth}
  \begin{picture}(1,0.528215)(0,-0.04)
    \put(0,0){\includegraphics[width=\FullWidth]{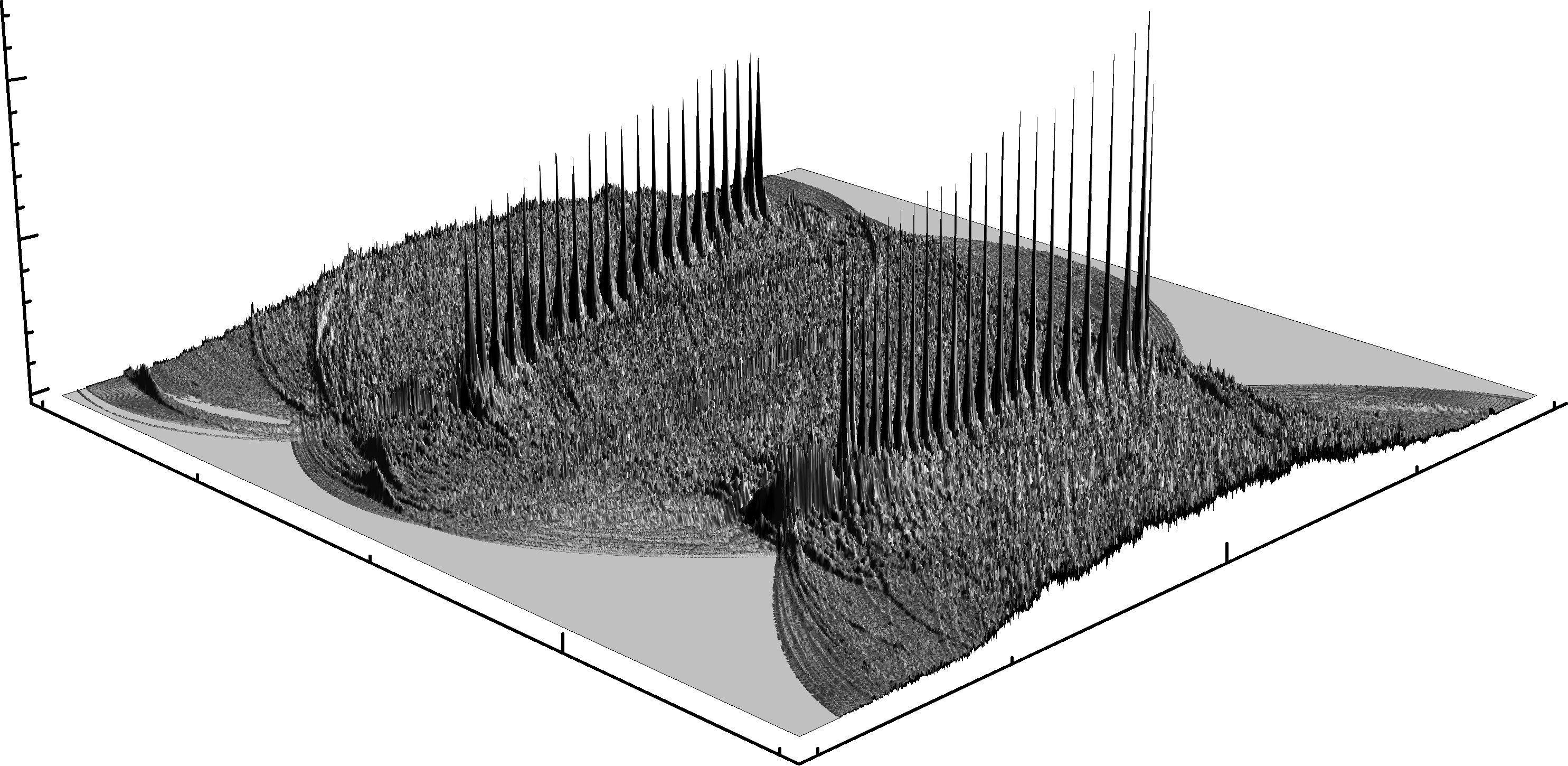}}
    \put(0.062172,0.444581){\makebox(0,0){\footnotesize{$400$}}}
    \put(0.069579,0.347610){\makebox(0,0){\footnotesize{$200$}}}
    \put(0,0.230355){\makebox(0,0){\footnotesize{$0$}}}
    \put(0.026936,0.187795){\makebox(0,0){\footnotesize{$\ang{135}$}}}
    \put(0.125589,0.141818){\makebox(0,0){\footnotesize{$\ang{90}$}}}
    \put(0.235690,0.089966){\makebox(0,0){\footnotesize{$\ang{45}$}}}
    \put(0.358586,0.032391){\makebox(0,0){\footnotesize{$\ang{0}$}}}
    \put(0.446970,-0.014106){\makebox(0,0){\footnotesize{$\ang{-45}$}}}
    \put(0.561212,-0.014106){\makebox(0,0){\footnotesize{$-\frac{\sqrt{13}}{2}$}}}
    \put(0.645118,0.025657){\makebox(0,0){\footnotesize{$-1$}}}
    \put(0.782492,0.089630){\makebox(0,0){\footnotesize{$0$}}}
    \put(0.903367,0.146869){\makebox(0,0){\footnotesize{$1$}}}
    \put(0.973985,0.177946){\makebox(0,0){\footnotesize{$\frac{\sqrt{13}}{2}$}}}
  \end{picture}
  \caption{Hough transform of the image in figure~\ref{fig:HT_DifferenceOfGaussians}. Each point in the $xy$-plane represents a line in the original picture: the $x$-axis is the distance of the line from the pole, the $y$-axis is the angle of incline of the line and the $z$-axis is its Hough value. The two sets of almost aligned peaks contain the lines of the grid (and the borders of the goban): in the set near the lower-right side there are the transverse lines (as seen in figure~\ref{fig:HT_DifferenceOfGaussians}), while longitudinal lines are included in the set near the upper-left side.}
  \label{fig:HT_HoughSpace}
\end{figure}
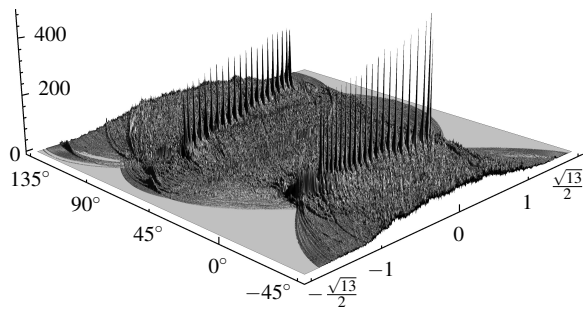

\SssSec{}\label{par:HT_MainHough} The Hough transform~\cite{DH72} of the B/W image is computed (see figure~\ref{fig:HT_HoughSpace}). Being the Hough transform a very stable and powerful tool to identify lines in a picture, it turned out that a complete and onerous calculation of the transform is not always required, especially for large photos. In order to boost performances, a complete calculation is carried out only for photos up to about half of a Mpixel; the larger the photo the smaller is the percentage of pixels we actually use to compute the Hough transform, reaching a minimum of 25\% for photos of 2.5 Mpixels (larger photos would be reduced). Pixels are homogeneously selected through a pseudo-random Richtmyer pattern~\cite{Ric58}.

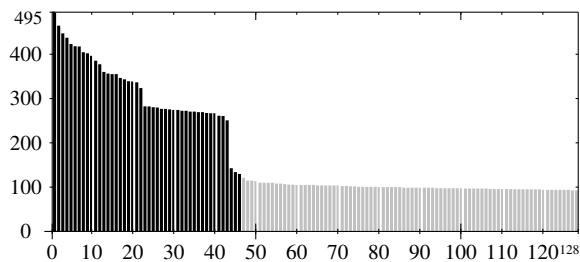
\begin{figure}[!htb]
  \centering
  \setlength{\unitlength}{0.927644\FullWidth}
  \begin{picture}(1.078,0.472)(-0.078,-0.052)
    \put(0,0){\includegraphics[width=0.927644\FullWidth,height=0.389610\FullWidth]{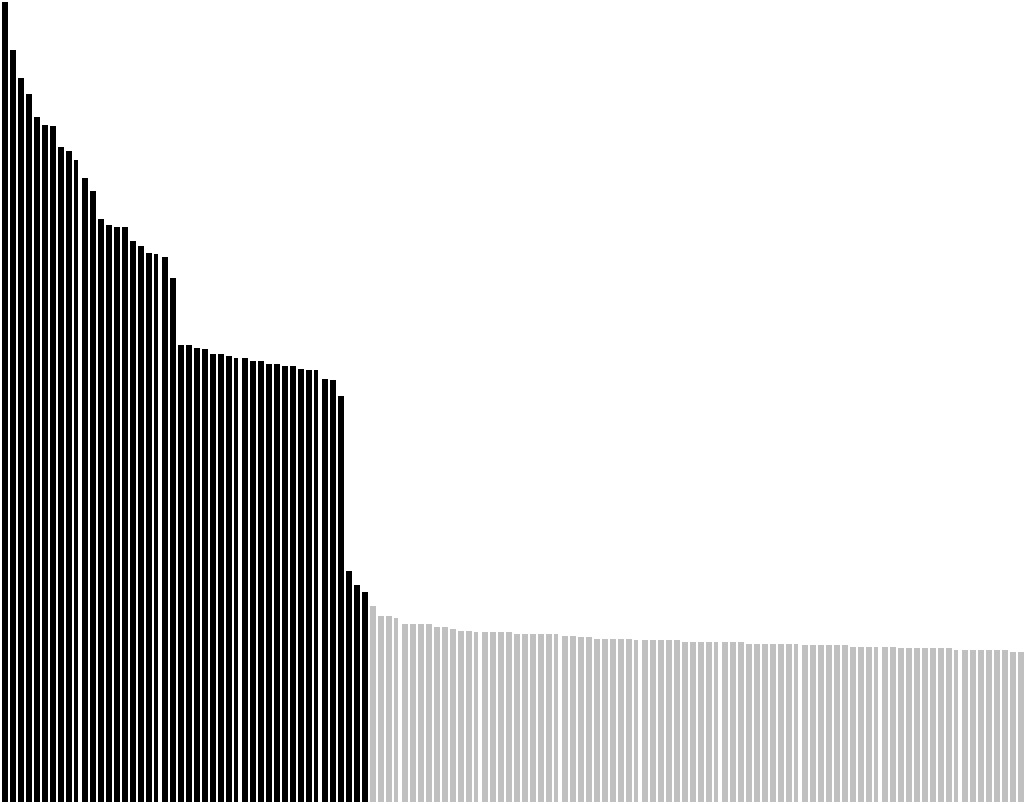}}
    \put(0,0){\line(1,0){1}}
    \put(0,0){\line(0,1){0.42}}
    \put(0,-0.005){\line(0,1){0.01}}
    \put(-0.005,0.42){\line(1,0){0.01}}
    \multiput(-0.005,0)(0,0.084848){5}{\line(1,0){0.01}}
    \multiput(0.995,0)(0,0.084848){5}{\line(1,0){0.005}}
    \put(-0.015,0){\line(1,0){0.03}}
    \put(-0.05,0){\makebox(0,0){\footnotesize{$0$}}}
    \put(-0.05,0.084848){\makebox(0,0){\footnotesize{$100$}}}
    \put(-0.05,0.169697){\makebox(0,0){\footnotesize{$200$}}}
    \put(-0.05,0.254545){\makebox(0,0){\footnotesize{$300$}}}
    \put(-0.05,0.339394){\makebox(0,0){\footnotesize{$400$}}}
    \put(-0.0448,0.4088){\makebox(0,0){\scriptsize{$495$}}}
    \put(0,-0.015){\line(0,1){0.03}}
    \multiput(0.075049,-0.005)(0.077973,0){12}{\line(0,1){0.01}}
    \multiput(0.075049,0.415)(0.077973,0){12}{\line(0,1){0.005}}
    \multiput(0.386941,-0.015)(0.389865,0){2}{\line(0,1){0.03}}
    \multiput(0.386941,0.405)(0.389865,0){2}{\line(0,1){0.015}}
    \put(1,-0.005){\line(0,1){0.01}}
    \put(0,-0.04){\makebox(0,0){\footnotesize{$0$}}}
    \put(0.075049,-0.04){\makebox(0,0){\footnotesize{$10$}}}
    \put(0.153022,-0.04){\makebox(0,0){\footnotesize{$20$}}}
    \put(0.230995,-0.04){\makebox(0,0){\footnotesize{$30$}}}
    \put(0.308968,-0.04){\makebox(0,0){\footnotesize{$40$}}}
    \put(0.386941,-0.04){\makebox(0,0){\footnotesize{$50$}}}
    \put(0.464914,-0.04){\makebox(0,0){\footnotesize{$60$}}}
    \put(0.542887,-0.04){\makebox(0,0){\footnotesize{$70$}}}
    \put(0.620860,-0.04){\makebox(0,0){\footnotesize{$80$}}}
    \put(0.698833,-0.04){\makebox(0,0){\footnotesize{$90$}}}
    \put(0.776806,-0.04){\makebox(0,0){\footnotesize{$100$}}}
    \put(0.854779,-0.04){\makebox(0,0){\footnotesize{$110$}}}
    \put(0.932752,-0.04){\makebox(0,0){\footnotesize{$120$}}}
    \put(0.9835,-0.0348){\makebox(0,0){\tiny{$128$}}}
    \put(0,0.42){\line(1,0){1}}
    \put(1,0){\line(0,1){0.42}}
  \end{picture}
  \caption{histogram of the $128$ highest local maxima found in the Hough transform shown in figure~\ref{fig:HT_HoughSpace}, sorted by decreasing values. The maxima that will be discarded as noise are drawn in grey, the intermediate plateau contains the longitudinal lines of the grid in figure~\ref{fig:HT_DifferenceOfGaussians} and, finally, the highest bars on the left include the transverse (and longest) lines of figure~\ref{fig:HT_DifferenceOfGaussians}. The boundary of the plateau has been given a wide tolerance to avoid any risk of throwing away useful data.}
  \label{fig:HT_MaximaHistogram}
\end{figure}

\SssSec{} The ``stronger'' is a line in the original image, the higher is its corresponding peak in the Hough space. Thus, the local maxima of the transformed image are singled out and sorted: assuming the goban is the main subject of the photo, its grid lines should be among the highest local maxima. So, in a histogram they should form a quite clear ``plateau'' as in figure~\ref{fig:HT_MaximaHistogram}, well above the background noise, which will be discarded in subsequent computations. \\ The sorting algorithm used is an adapted version of bubble sort: the values to be sorted may be several thousands, yet the computational complexity is linear as we are only interested in a quite small and preset amount of the highest local maxima.

\SssSec{} The grid of any goban is formed by two perpendicular sets of parallel lines. Taking a photo means to project those lines into the plane of the camera sensor: if the lines are parallel to the sensor their projection is a set of parallel lines too, otherwise they are projected into a set of converging lines. When in step~\ref{par:HT_MainHough} those sets of lines are transformed into points of the Hough space, they lie either on a horizontal straight line (former case) or in a sinusoid (latter case). Under at least fair conditions, which we define in section~\ref{sec:Conclusion}, the relevant stretch of the sinusoid is either very small, compared with its amplitude, or near its inflexion point, or both, thus a very good approximation of a straight line. That justifies the use of a second run of the Hough transform (provided a suitable tolerance) to identify \textit{in the first Hough space} the two sets of lines of the grid. As the local maxima selected in the previous step are at most a few dozens of points, that second transform is fast and more accurate than RANSAC or similar algorithms.

\SssSec{} Once the two mutually orthogonal sets of parallel lines are identified, they are pruned of spurious lines (e.g.: the wooden borders of the goban). This task is accomplished by evaluating the median distance between the lines of the set and excluding those lines whose placement is worse fitting, even when taking the effects of projections into account.

\SssSec{}\label{par:HT_Interpolation} The same above-mentioned expected median distance is used to interpolate missing lines as well: if there is a gap in a set of parallel lines, caused by any contingent disturbance in the photo, and if there is at least a ``low'' local maximum in the Hough transform where one or more lines are missing, then those lines are inserted in the set.

\SssSec{} Among the lines selected in the previous steps an attempt is made to select the subset of $19+19$ lines most likely forming the grid, minimizing a norm that measures the uniformity of the distances between the lines. In case of failure another attempt is made with $13+13$ lines only, then $9+9$. In the highly infrequent event that last attempt fails too, the whole procedure is aborted and the user has to set the size of the goban and to manually pinpoint the four corners of the grid in the photo.

\SssSec{} If the grid has been found, the lines forming it are mutually intersected to compute the placement of each grid point. Furthermore, the actual spacings between lines are recorded for future use, this being the rationale: the coordinates of the internal grid points could be geometrically computed by knowing the coordinates of the four corners alone, but it happens the grids of most gobans are not perfectly drawn. Therefore, using the recorded spacings instead of the equidistant ones leads to a more precise correspondence between the actual grid points in the photo and the computed ones.

\subsection{Automatic micro-recalibration}\setcounter{SssSec}{0}

The grid of the goban must be accurately pinpointed for each and every photo of the game, yet applying every time the algorithm discussed in subsection~\ref{subsec:StartingGrid} is not an option, for both practical and theoretical reasons. The former is the sheer computational complexity: even on high-end up-to-date personal computers it requires from one to a few seconds to be executed (most of the time is spent to calculate the Hough transform, depending on size, digital noise and other features of the analysed photo). The latter reason is the fact that the more the game progresses, the more the stones conceal the grid lines: experiments show that after 150 moves or so that algorithm rarely succeeds. On the other hand we can not suppose that the grid always remains in the same position inside the photos throughout the entire game: even if a stand is used, vibrations may occur, players may hit the camera, or the stand, or the goban and so on.

Provided the corners of the grid are not too close to the edges of the photos (a minimum distance of approximately the diameter of a stone is required), three different variations of the same idea successfully solve the issue almost every time. First of all we assume that, if a movement really occurs between two consecutive photos, the displacement of the image of the grid between those photos is small (up to the radius of a stone). This assumption allows faster calculations and it is almost automatically fulfilled in case of a program dealing with video streams, as suggested in section~\ref{sec:Conclusion}. Then, starting from the first one, the procedures described below are applied, each one only after the failure of the previous one: if one step succeeds the search ends there --- when they all fail the entire procedure described in subsection~\ref{subsec:StartingGrid} must be run anew. In this way we generally attain the tracking of small movements of the grid within a time in the order of magnitude of a tenth of a second.

\SssSec{Corner Hough transform}\\
In a neighbourhood of the last known position of each corner of the grid we compute an adapted version of the Hough transform, modified to detect corners whose sides are nearly parallel to the external lines of the latest recognised grid (this is coherent with the small displacement hypothesis). Using techniques derived from projective geometry, we are able to discard false positives or to detect the new position of the grid even if one corner is not recognized or it is hidden by a stone: to accomplish that we compute the complex cross-ratio of the four corners of the grid, using their $x$ and $y$ coordinates in the photo as real and imaginary part respectively of a single complex number. That complex cross-ratio is preserved by rotations, translations and homogeneous dilations, but it is not preserved by \textit{real} central projections, providing an useful tool for detecting substantial changes in the point of view or errors in the pinpointing of the corners. \\ This method should work almost to the end of the game, when two or more corners may be covered by stones, resulting not visible in the photos.

\SssSec{Linear Hough transform}\\
A fragment of the (standard linear) Hough transform is computed in a neighbourhood of the latest known position of each external line of the grid --- actually only around those segments of the lines where we know there are no stones (but the last played and thence not yet detected one). \\ As the previous method, this one should work up to the last moves of the game too, when a side of the grid may happen to be almost entirely concealed by stones. For that reason, in order to further improve time performance (with a negligible worsening of success rate), it may eventually be discarded.

\SssSec{Elliptic Hough transform}\\
A fragment of the Hough transform for elliptic shapes is calculated in a neighbourhood of each known stone placed in the external lines of the grid; the positions of the lines are thereby deduced through a linear regression applied to the coordinates of the centres of the recognised stones, taking into account both projective distortions and systematic misplacements of the stones. \\ This method is not applicable in the first part of the game, a phase when usually there are almost no stones in the external lines, but it becomes more and more accurate towards the end of the game.

\subsection{Approximation of the point of view}

\begin{figure}[!htb]
  \centering
  \includegraphics[width=\FullWidth]{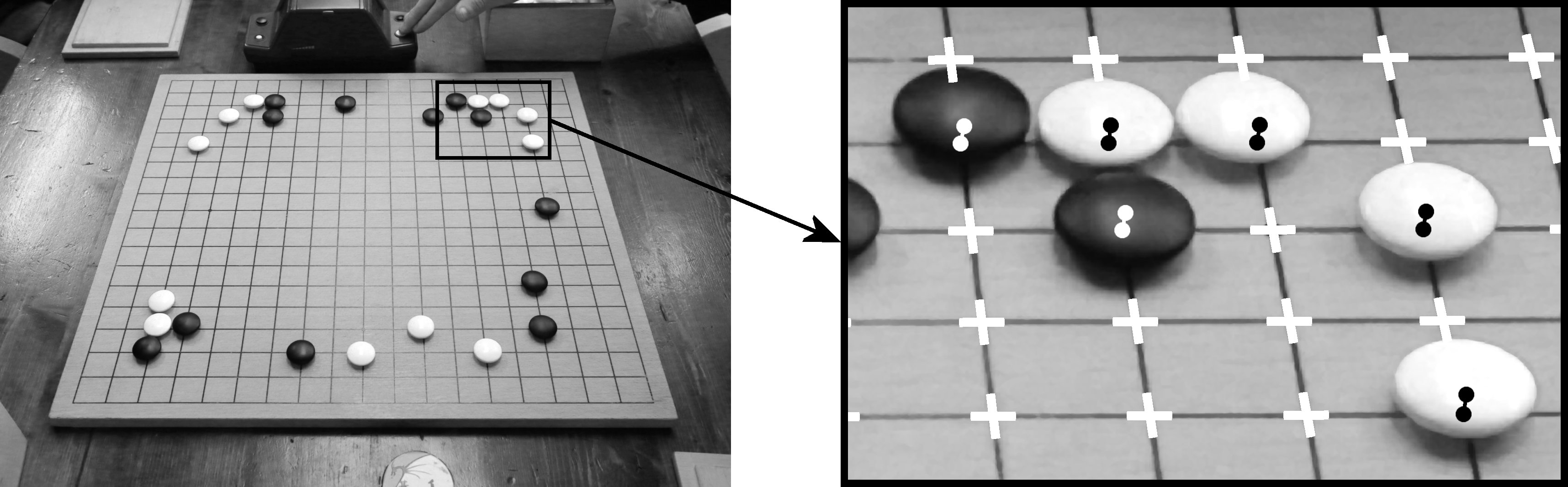}
  \caption{discrepancy between the computed grid points (the white crosses where there are no stones, the lower dot where there are) and the computed projection of the geometrical centre of the stones (the upper dot), the latter clearly being a better approximation of the actual location of the stones, even the misplaced ones.}
  \label{fig:PoV_Discrepancy}
\end{figure}

Even if a stone is perfectly placed upon a grid point, which hardly ever happens (but can be assumed as the mean placement), the projection of its geometrical centre does not coincide with the grid point in the photo, unless the camera is on the ``point at infinity'' of the normal line to the plane of the goban, which obviously never happens. The discrepancy is substantial for the stones in the furthest lines of the grid, especially if the photos are taken from a low point of view, nevertheless it is non-negligible even in photos shot under good conditions, as shown in figure~\ref{fig:PoV_Discrepancy}.

Being the goban flat, the knowledge of its position in the photo gives no information about the third dimension and the orientation of the $z$-axis in each grid point. In order to obtain those information it is necessary to approximate the position and orientation of the camera actually taking the photos and to use those data to calculate the projection into the plane of the image sensor of any point of the space which we are interested in. The problem of finding the point of view is typically resolvable when two or more pictures of the same scene are available, for instance using techniques of epipolar geometry, or when it is possible to implement an interactive search, as shown in~\cite{BAD10}. Unfortunately, none of those approach are applicable in our case, as we need the point of view for each photo separately. Furthermore, according to Hadamard's definition, the general problem of finding the point of view of a single image is ill-posed.

What we can exploit to solve it is the a priori knowledge that we are observing the projection of an approximately square-shaped object and that we are indeed only interested in shapes, not in absolute dimensions (i.e.: we do not need to know how big the goban is or how distant it is from the camera). So we set a relative unit of measure (half of the longitudinal side of the grid, as seen by the camera) and an origin for the Cartesian coordinate system (the ``tengen''\footnote{~Tengen is a Japanese Go term literally meaning ``origin of heaven'': it denotes the central point of the grid.}), with $x$-axis and $y$-axis parallel to the lines of the grid and $z$-axis pointing upwards. Like any rigid body, the camera has six degrees of freedom: the three coordinates of the nodal point $N$ of the lens (which is the actual point of view), the orientation of the camera (which is defined by the two coordinates of the point $G$ at which the camera is aimed in the 
plane of the goban) and finally the angle of roll of the camera along the line $NG$ (which is zero if the camera is level, non-zero if the camera is leaning). Direct geometrical calculations allow to obtain the vanishing points of the lines and diagonals of the grid, which in turn are used to compute the horizon and then the angle of roll of the camera. Moreover, the coordinates of $G$ are evaluated by applying real cross-ratios and, finally, it is possible to infer the equation of the vertical plane $\nu$ containing $N$ from the intersections among the lines of the grid and the line perpendicular to the horizon passing through $G$.

\begin{figure}[!htb]
  \centering
  \includegraphics[width=\FullWidth]{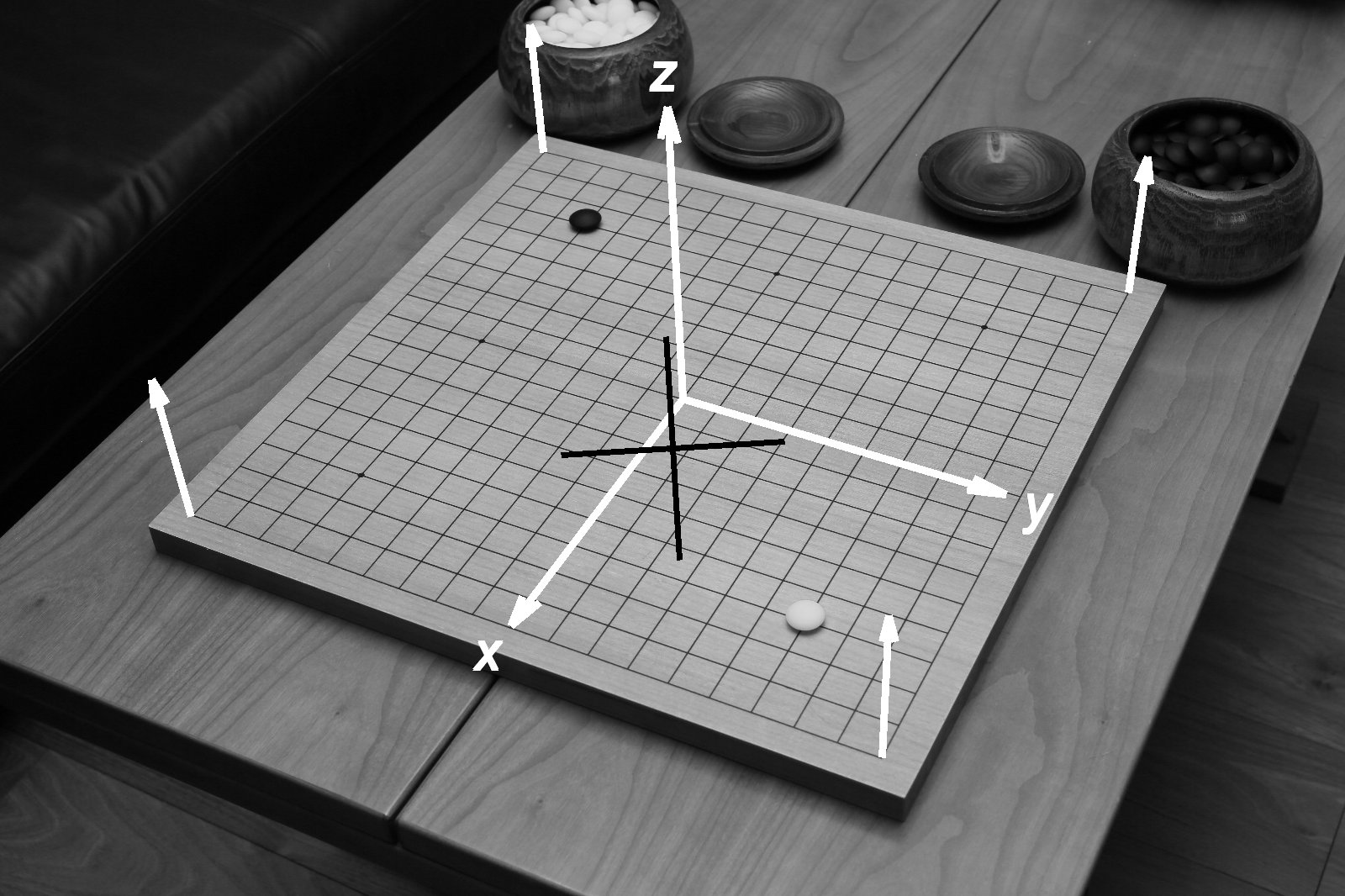}
  \caption{the leaning central black cross marks the point where the camera is aimed at; computed values for this photo are: $N(3.789,2.072,3.667)$, $G(0.207,0.072)$, $\text{angle of roll}=\ang{-3.34}$ and $t=1.038$ (Chinese goban placed transversely). The three-dimensional orthonormal basis is drawn in white, as well as the vertical half-versors in each corner of the grid.}
  \label{fig:PoV_Axes}
\end{figure}

That completes the list of quantities immediately and explicitly deducible from the analysis of the photo. Two more coordinates remain to be approximated: those of $N$ in $\nu$, for which we use an algorithm derived from both the shooting method and the bisection one used in numerical analysis.
\begin{enumerate}
  \item Apart from some handmade amateurish gobans, for aesthetic reasons the grid is not square: to compensate for perspective effects the ratio between its sides is about 1.077 (for Japanese and Korean gobans) or 1.038 (for Chinese gobans). Since the dilation along a single axis is not a projection and since the longest side of the grid may be either the longitudinal or the transverse one in the photo, repeat steps from~\ref{alg:PoV_Cicle} to~\ref{alg:PoV_Norm} for $t\in\{1.077,\frac{1}{1.077},1.038,\frac{1}{1.038},1\}$.
  \item Let $\varphi_0=\ang{45}$ the angular coordinate of a point $N_0$ in the polar coordinate system of $\nu$ with pole $G$ and horizontal polar axis; let $\rho_0=5$ its radial coordinate (with the same unit of measure of the main Cartesian system, thus 5 units are equivalent to 2.5 times the length of the longitudinal lines of the grid).\label{alg:PoV_Cicle}
  \item Starting with $k=0$ and using $N_k$ as point of view, project a rectangle whose corners are the points $(\pm{1},\pm{t},0)$ into the plane containing $G$ and perpendicular to $NG$, thus obtaining a projection homothetic to the one lying in the plane of the image sensor --- whose location is unknowable.
  \item Slightly altering the values of $\rho_k$ and $\varphi_k$, evaluate the projective convergence of opposite longitudinal sides of the grid and the angle between the main diagonals. Hence, comparing those angles to the actual ones in the photo, for increasing values of $k\in\mathbb{N}$ accordingly set the new values of the couple \[
    \begin{cases}
      \rho_{k+1} = \rho_k+\Delta_{\rho_k} \\
      \varphi_{k+1} = \varphi_k+\Delta_{\varphi_k}
    \end{cases}
  \] and iterate this and the previous step until $N_k$ converges (with quickly decreasing values for $\Delta_{\rho_k}$ and $\Delta_{\varphi_k}$ convergence is guaranteed).
  \item Uniformly scale the last computed grid to best fit the one in the photo and calculate the maximum distance between each corner of the grid and the actual corners found in the photo.\label{alg:PoV_Norm}
  \item Chose the value of $t$ that minimises the distances of step~\ref{alg:PoV_Norm}, thus defining the most likely kind and orientation of the goban and --- as a by-product --- the point of view (see figure~\ref{fig:PoV_Axes}).
\end{enumerate}

\section{Detecting the stones}
\label{sec:DetectingStones}

\subsection{From theory to practice}

\begin{figure}[!htb]
  \begin{minipage}[t]{\HalfWidth}
    \includegraphics[width=\textwidth]{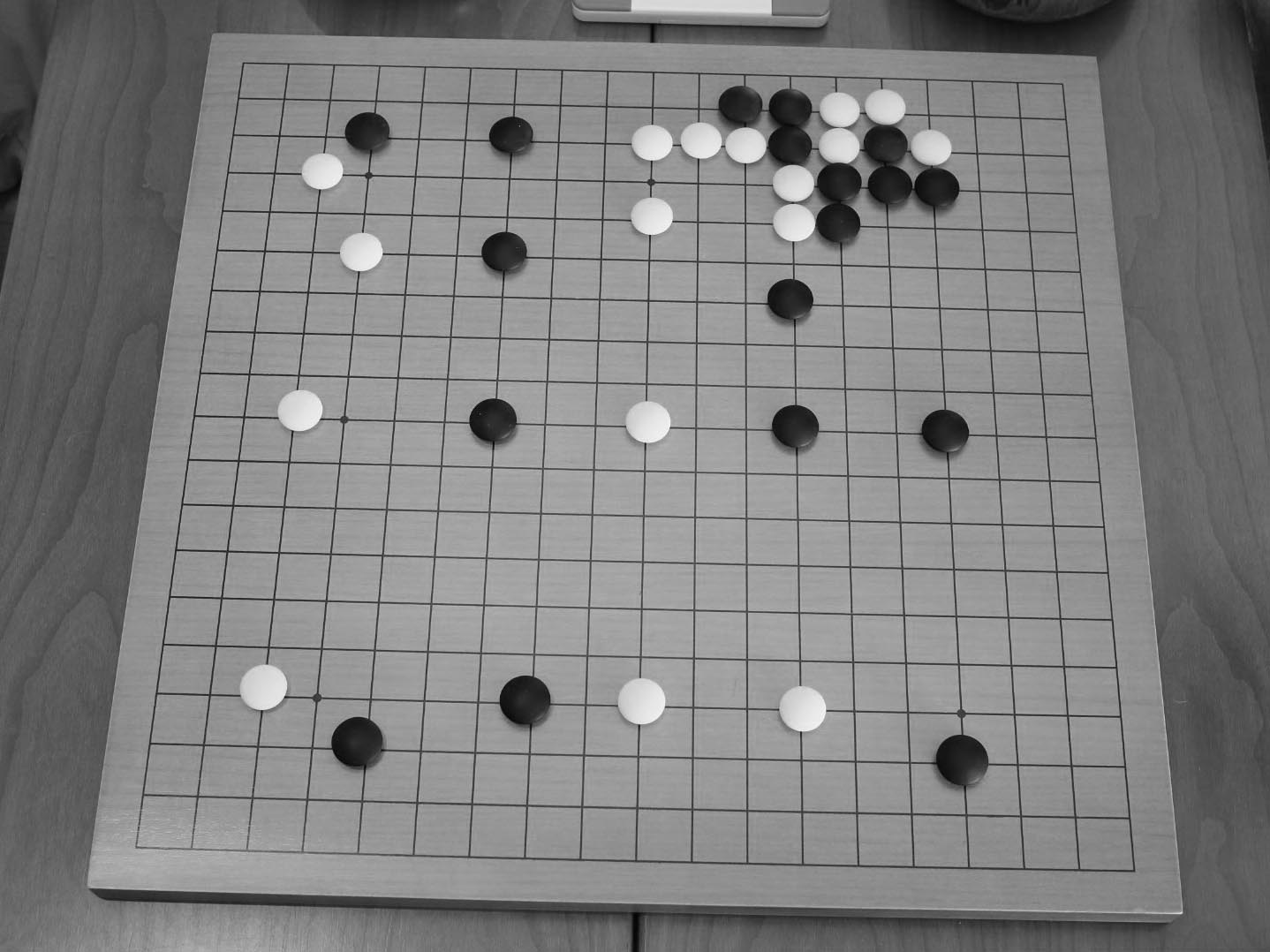}
    \caption{good conditions.}
    \label{fig:optimal}
  \end{minipage}
  \quad
  \begin{minipage}[t]{\HalfWidth}
    \includegraphics[width=\textwidth]{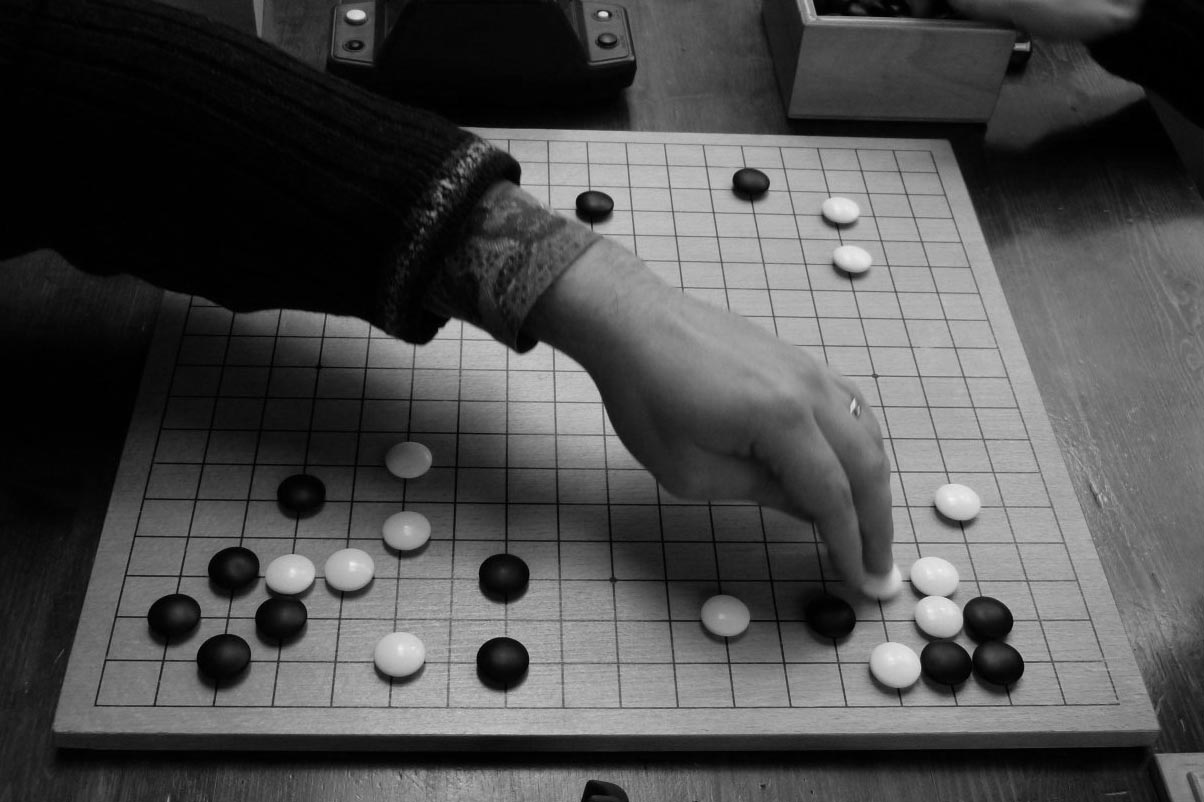}
    \caption{poor conditions (low point of view, player's arm visible).}
    \label{fig:notoptimal}
  \end{minipage}
\end{figure}

As previously pointed out, detecting stones on a goban is not a difficult task given optimal conditions, but real games are another matter and a lot of problems invariably arise. We have already discussed what happens when the table is bumped or the camera vibrates, forcing an automatic recalibration of the grid, but most issues emerge because good conditions, such as in figure~\ref{fig:optimal}, are uncommon. In figure~\ref{fig:notoptimal} we see instead which conditions are typical of real games,
during which there is no control over the point of view --- standing too close would disturb the players --- neither on the light, both factors being crucial for the detection process. The players themselves, who usually pay little care to the camera, may cause further problems, making it difficult to take pictures without some fingers, hands and even whole arms covering many, or most stones. Also, in the heat of the game it's not uncommon to skip some pictures or, on the contrary, to take more than one of a single move.
\\Fixing these latter mistakes is easy (for example, if a photograph is skipped the last stone put on the goban will be certainly missed, but, together with the next one, could be detected anyhow in the following picture), but the real challenge is detecting the stones under poor conditions, such as the ones depicted above. Unless error-proof algorithms were employed, such conditions could result in a frustrating series of errors that at best would transform the analysis in a slow, painful process, and at worst could even go unnoticed, turning the analysis into a disaster, with a likely wrong score.
\\To clarify the extent of such problems, let's just mention that during the ``Il David'' Tournament, held in Florence in December 2012, we recorded a game by means of 269 photographs, each one taken after a single move. Of these photographs, 65 --- 1 out of 4, and up to 6 consecutive --- were affected by ``disturbance'', such as fingers/hands/arms of the players. Also, dark shadows were often cast on the goban, further affecting stone detection. Without error-proof algorithms, capable of detecting the stones no matter what, any automatic reconstruction of such a game would inevitably fail and, instead of losing a lot of time trying to get over the continual errors thrown by simple algorithms, it would be better to switch to a manual inspection of the pictures, in order to at least reconstruct the game by hand.

\subsection{Detecting all stones vs. detecting last stone}\setcounter{SssSec}{0}

Before choosing an algorithm, it's important to decide which one of the following two approaches is the best: the ``traditional'', implemented by \textit{Kifu Snap} and \textit{Imago}, for example, that is starting from scratch on every picture in order to locate the grid and every single stone on the goban, or the ``minimal'', that is relying heavily upon data collected from previous pictures and trying to detect only the last stone put on the goban.
We found the second approach to be the best one, at least when the goal is the reconstruction of a whole game and not just the identification of a specific position. There are at least four reasons to support such a choice:

\SssSec{} Just by looking at figure~\ref{fig:notoptimal} it is clear there is no way to guess which stones may be hidden by the ``disturbance'' (a whole arm), no matter how good the algorithm employed is. The ``traditional'' approach would always fail, while the ``minimal'' one, were the last stone put on the goban still visible, would likely succeed, at least in theory.

\SssSec{} The second approach, having just one stone to detect, is much faster than the first one. This is not immediately clear, as every intersection has to be checked, no matter how many stones have to be found. But as the second approach relies on data collected from previous pictures, it knows where the empty intersections are and may limit its searching to them, thus reducing the time required by the calculations; also, knowing for certain that all the intersections but one are empty, it can examine thoroughly only the most promising ones, discarding the others after a rough examination. That's why \textit{Imago}, the program mentioned before, takes about 28 seconds per move (although most of the time is needed for locating the grid), as it makes use of the first, ``traditional'' approach. The second, ``minimal'' approach, could require only a fraction of second per move.

\SssSec{} Although not so obvious at first, looking for all the stones is more error-prone than looking for just one. For example, if the first approach employs an algorithm with a known success rate of 99\% --- meaning that, for every intersection, it will correctly guess 99\% of the times if it's empty or if a stone is there --- it would seem, at first, that the outcome will be very good. But on closer scrutiny it becomes clear it is by no means good: as there are 361 intersections on a standard goban, the algorithm will fail, on average, on 3.61 of them and for each picture the chance of turning out error-free will be just $0.99^{361}$ = 2.7\%. This means that almost all pictures in a game will be affected by errors, most of the times multiple ones, and no automatic reconstruction will ever be possible. Even a 99.9\% success rate would not suffice, as just about 70\% of pictures will be error-free, a percentage not high enough.\!\footnote{~It requires an unbelievable 99.99\% success rate in order to lower the picture failure rate to a reasonable 3.5\%.} That's why \textit{Imago}, despite claiming a remarkable success rate of 99.86\%, fails on 24\% of its test pictures (13 out of 55).\!\footnote{~That's better than the expected percentage of 40\% because the test pictures were not chosen randomly, resulting in a more compact error distribution.}
\\If we make use of the ``minimal'' approach instead, we could employ two algorithms, as hinted before: one to select the most promising intersections, the other to analyse in depth these last ones. Let's now assume the second algorithm could claim, once again, a success rate of 99\% --- being possibly the same algorithm employed before ---; let's also assume ten or so stones must be selected to allow the first algorithm an equal success rate of 99\% --- meaning now that 99\% of the times the stone we're looking for will indeed be found on one of these intersections. If the assumptions were correct, the chance of each picture of turning out error-free will be $0.99^{11}$ = almost 90\%: a very good result despite the same success rates proved inadequate when making use of the first approach. Furthermore, according to our tests only two or three intersections have usually to be selected to let the first algorithm achieve a success rate of 99\%: selecting ten will further increase the overall efficiency of the ``minimal'' approach.

\begin{figure}[!htb]
\centering
  \includegraphics[width=\FullWidth]{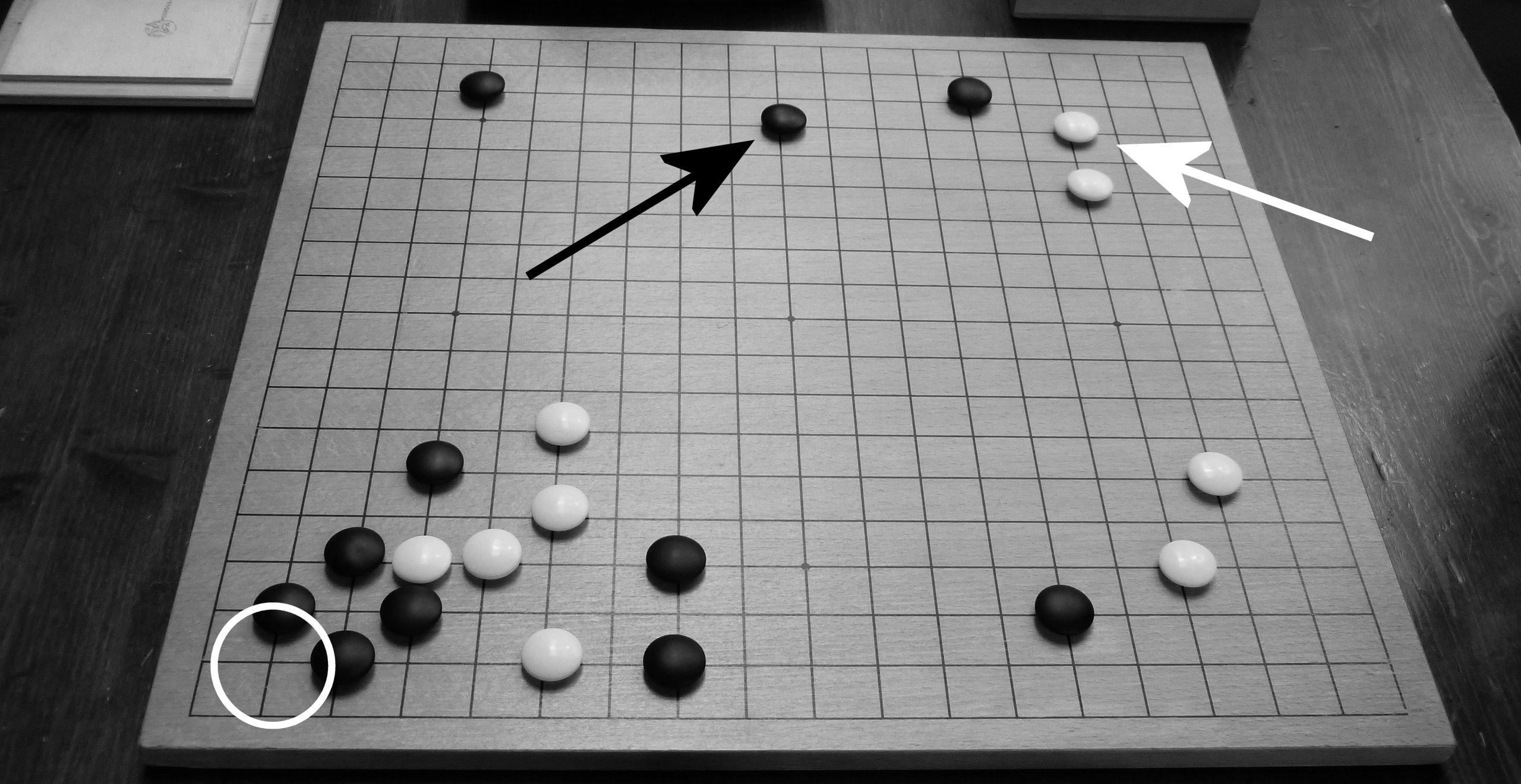}
  \caption{}
  \label{fig:thrNOOK}
\end{figure}

\SssSec{} The ``minimal'' approach makes things easier, because locating the stones and avoiding false negatives/positives is much simpler. For example, a simple way of locating the stones is to compute the luminance of the intersections, relying on the likely equivalence: high luminance $\Rightarrow$ a white stone; low luminance $\Rightarrow$ a black stone; in the middle, it's empty. A simple procedure could be:
\begin{itemize}[leftmargin=*]
\setlength\itemsep{0em}
  \item computing, for every intersection, the value of its luminance, averaging that of many pixels around the centre;
  \item establishing a ``high threshold'' and a ``low threshold'' by means of test pictures;
  \item if \texttt{luminance of the intersection~>~high threshold}, a white stone is there; if \texttt{luminance~<~low threshold}, a black stone is there; if \texttt{luminance in between}, it's empty.
\end{itemize}
This procedure usually works well enough. But nothing guarantees these same thresholds will work on the next pictures, as a small change in the light could easily alter the luminance distribution. And in other games the situation could get even worse. For example, the picture in figure~\ref{fig:optimal} has an `high threshold'' at 215 (out of 255, as the values are expressed by means of unsigned bytes) and a ``low threshold'' at 100, but if the same values are applied on the picture in figure~\ref{fig:thrNOOK}
we'll miss, albeit by a small margin, the stone pointed out by the white arrow (a false negative) and mistake the empty intersection marked by the circle for a black stone (a false positive).
\\Improving the algorithm does not work: let's try, for example, to dynamically set the thresholds, computing first the mean luminance of the whole goban, then adding/subtracting a fixed percentage. In such a case the picture in figure~\ref{fig:optimal} needs a ``high threshold'' of 33\% (that is, the luminance of white stones is always greater than the mean luminance of whole goban + 33\%), that is not enough to detect, again, the stone pointed out by the white arrow in figure~\ref{fig:thrNOOK}. The two luminance distributions are simply too much apart.
\\To solve the problem once and for all, two things are needed: first, finding more criteria to combine together; second, restricting the analysis to the empty intersections only, in order to get rid of the misleading contribution of the stones. And that's why the ``minimal'' approach is needed.
\\In order to better understand this point, let's have a look at the luminance distribution of the picture in figure~\ref{fig:thrNOOK},
\begin{figure}[!htb]
  \centering
   \includegraphics[width=\FullWidth]{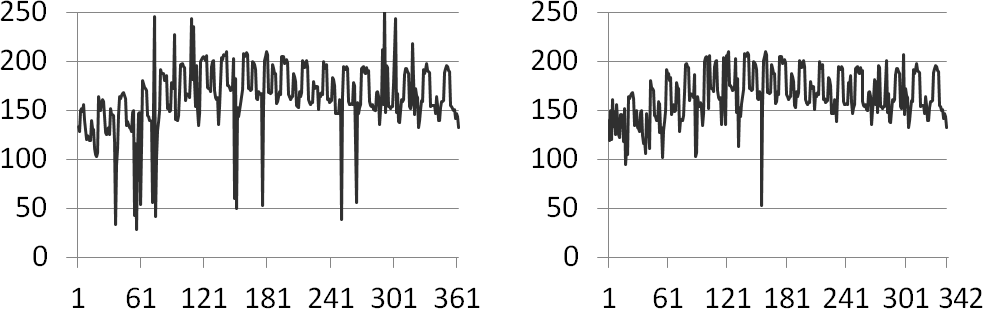}
   \caption{luminance distribution of figure~\ref{fig:thrNOOK} intersections.}
   \label{fig:approaches}
\end{figure}
first considering all the intersections, then removing all the stones except for the last one put on the goban: it's quite obvious that the single peak on the right, which matches the last stone put on the goban (pointed by the black arrow in figure~\ref{fig:thrNOOK}), is much easier to find than all the peaks on the left. It would suffice to pick the smallest value of the distribution function to find the stone, something not feasible on the other one, where the same peak is not the most prominent. And although most peaks may easily be identified, some may not --- the drawback of having so many to deal with --- and require further analysis to rule out any false positive and, of course, not to miss any stone.

\subsection{Choosing an algorithm}

After having chosen the ``minimal'' approach (searching for the last stone only) over the ``traditional'' one (searching for all the stones), we'll now describe which criteria are best suited for finding the stones and why, then how to combine them, and eventually which use to make of the ``combination''.

\subsubsection{The criteria}
\label{sssec:Criteria}

\begin{figure}[!htb]
  \centering
  \includegraphics[width=\FullWidth]{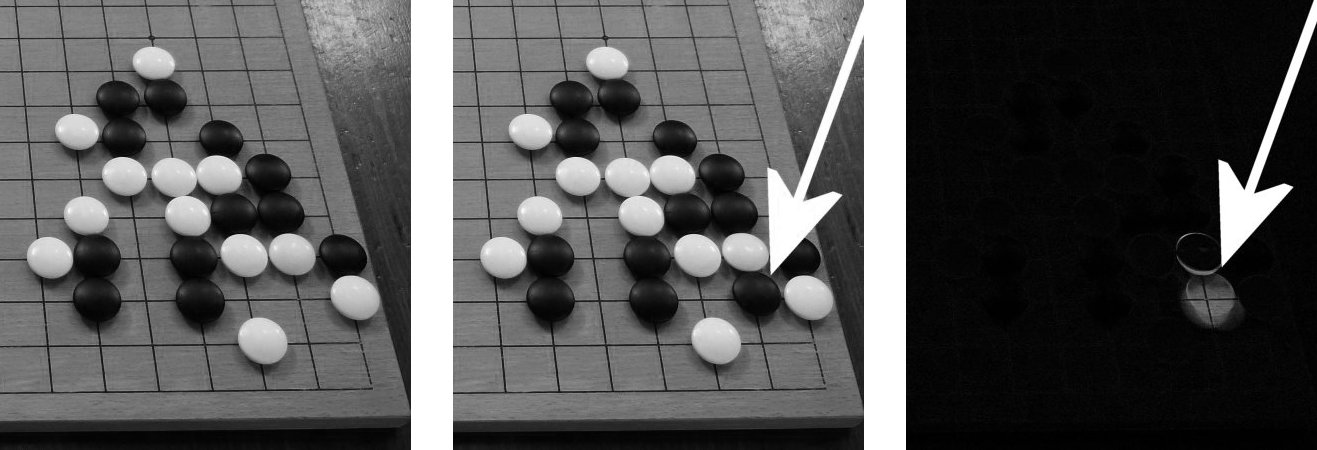}
  \caption{difference between consecutive pictures: only the last stone put on the goban stands out clearly.}
  \label{fig:difference}
\end{figure}

\SssSec{Difference between pictures}
\\This criterion is simple: without ``disturbance'', two consecutive pictures of the same game should be identical except for the last stone put on the goban.\!\footnote{~The difference would also extend to the stones possibly captured but, in such a case, could be neglected, as the ``minimal'' approach only evaluates empty intersections.} Computing the difference between a picture and the previous one should immediately highlight the last stone only.
In figure~\ref{fig:difference} we see an example of such an operation, with a really good outcome.\\
\ProsCons{it's a fast operation and it's very easy to implement. In absence of disturbance it works very well, with reliable outcomes.}{it's extremely sensitive to disturbance, albeit small; furthermore, as any disturbance will also affect the difference with the next picture, it will be necessary to choose the lesser evil: either dealing with another dubious outcome or discarding the picture entirely, then losing valuable information.}

\SssSec{Analysis of local features}\label{par:Features}
\\We've already discussed the luminance, which is one of these features, albeit not a very useful one, as it's sensitive not only to disturbance, but also to changes in the light that could impose on the thresholds employed. Another feature is the so-called ``chrominance'', that is the standard deviation of the RGB components around an intersection: as the stones are white or black while the goban surface is usually yellowish, computing an intersection's chrominance would tell if it is colourful --- then empty --- or some shade of grey --- then likely white or black, thus covered by a stone. With a truly yellowish goban, chrominance analysis is very reliable, as shadows and reflections would never turn a colourful intersection into a greyish one; but this could happen anyhow under certain lights, making chrominance completely worthless, or even detrimental. Also, any major disturbance would have the same effect.
\\Other potential features are hue and saturation (first two components of the HSL colour system), both fully capable of highlighting the stones; furthermore, the hue is insensitive to shadows, even dark ones, while saturation is not affected by reflections. Unfortunately the opposite is also true: hue highlights any reflections, even faint ones, while saturation is sensitive to shadows.
\\The last feature is the so-called ``uniformity''. While the stones' surface look uniform, the goban's does not, because the intersections are crossed by the grid lines, which produce a local disturbance that disrupts the surface uniformity. A careful research of such small disturbance could possibly tell apart the empty intersections from the stones and even circumvent major disturbance. Problems arise on the borders, especially the corners, where some lines are missing, and in some kind of disturbance (dark, uniform sleeves, for example).\\
\ProsCons{all the features are easy to compute and, under good conditions, produce good outcomes. Some are also fast to compute.}{all the features are sensitive to disturbance and cannot be trusted alone; they need to be combined in a complex formula. Some are not fast to compute (uniformity, chrominance).}

\SssSec{Circular/elliptical Hough transform}\label{par:Hough}
\\The Hough transform, as previously discussed, is a mathematical process capable of highlighting specific features, such as lines, circles and so on, if present in the pictures. It could easily be employed for locating the stones, which look like small ellipses scattered on a flat surface. The process creates ``signals'' in the Hough space, each one matching one of the features we're looking for:
\begin{figure}[!htb]
  \centering
  \includegraphics[width=\FullWidth]{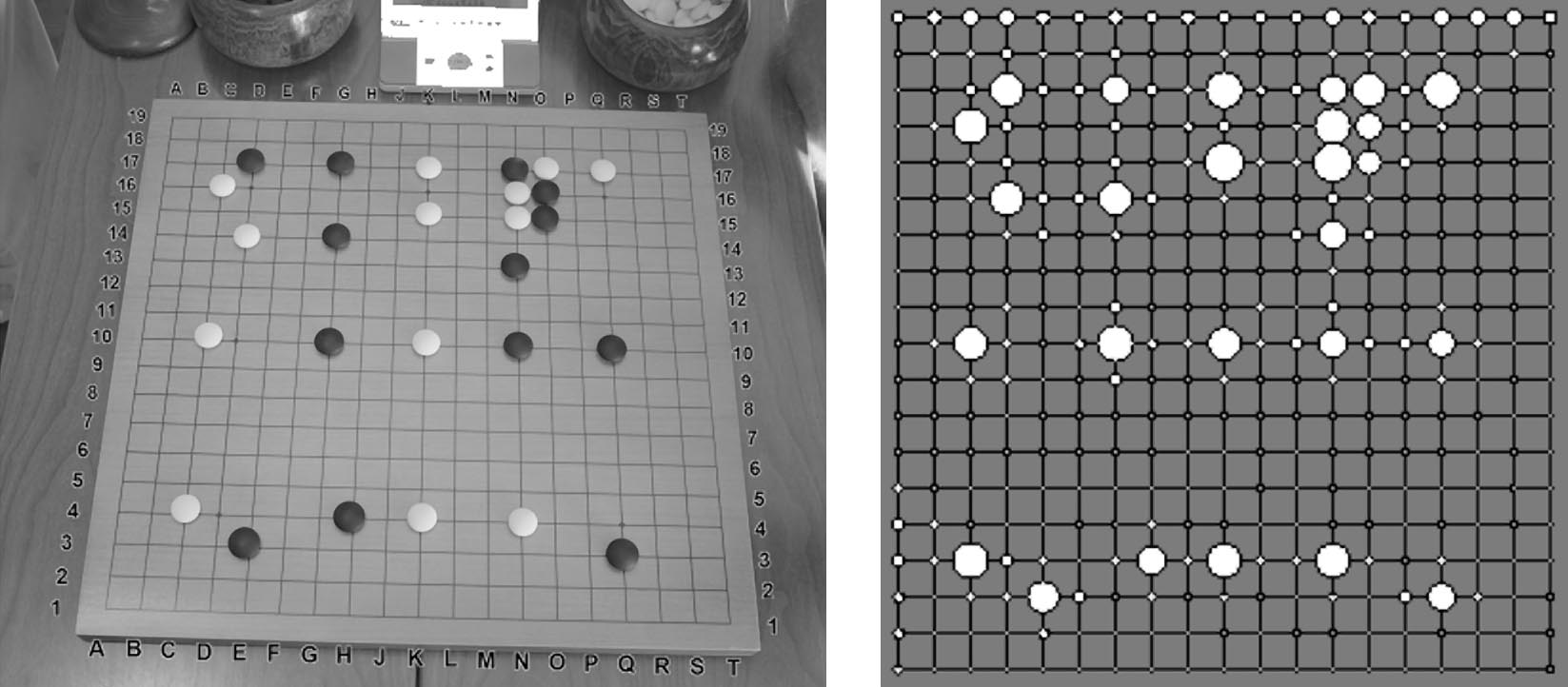}
  \caption{the outcome of the circular transform with a good point of view: strong signals matching the stones, and the stones only.}
  \label{fig:HoughOK}
\end{figure}
a graphical depiction of these signals is shown in figure~\ref{fig:HoughOK}, with the picture to inspect on the left and the relative Hough space on the right, after the application of the circular transform: the stronger a signal (the white circles) the more likely a stone will be found over the matching intersection.
\\At first, it looks like this process could solve the stones' detection problem once and for all, as the signals are strong where a stone is indeed placed over an intersection, weak otherwise, and even major disturbance, unless resembling small circles, won't affect the outcome. But on closer scrutiny figure~\ref{fig:HoughOK} shows that some of the weak signals are not so weak after all: for example, there are many on the goban's upper side. This is not a nuance: as the stones look more elliptical than circular, the further they are from the observer or the lower the point of view, the more difficult it is for the transform to detect them. In order to bypass the problem, an elliptical transform could be employed, but the time required would grow, becoming unsuitable for real-time analysis. Furthermore, the outcome of the transform is influenced by nearby stones, especially off-centre ones; some disturbance (for example fingertips) and even shadows cast by the stones themselves could also alter the signals.\\
\ProsCons{with a good point of view the outcome is always good, almost insensitive to disturbance.}{it's a slow process; if the point of view is bad the outcome becomes erratic; sensitive to some kind of disturbance not affecting other criteria.}

\subsubsection{The algorithm}

After a careful evaluation of the criteria pros and cons, we gave up the slow Hough transform, instead relying heavily upon the ``difference between pictures''. We also made use of all the local features discussed before and step by step the success rate improved, eventually reaching a limit when the ``short blanket'' effect appeared: any variation of the thresholds that could, for example, decrease the number of false negatives, made new false positives arise, and vice versa. A 100\% success rate was indeed achieved, but only under good conditions and no disturbance: under fair conditions it did not exceed 98-99\% and, if disturbance was present, never went beyond 25-30\%. This was by all means a remarkable feat, but there was no real improvement over the past attempts, the ones we listed in the introduction: the advantage of the ``minimal'' approach was clear, still not decisive.
\\A breakthrough was eventually made when a friend of us, the renowned amateur mathematician Dani Ferrari, M.Sc., especially known for his work on alphametics, recommended a new technique in order to make the most of the ``minimal'' approach. He observed that most of the times it was not possible to tell apart disturbance from stones and suggested to examine all the intersections, not just the empty ones, in order to gather information about ``how the stones look like''. This meant sorting the intersections in three groups, empty, covered by a white stone, covered by a black stone (using data from previous pictures), and computing the local features' mean values for each one of them; after that, a comparison between these mean values and those of an empty intersection should have disclosed the nature of the latter. A small difference with the ``mean white stone'', for example, together with a big difference with the ``mean empty intersection'' would have meant that a white stone was over the intersection under scrutiny.
\\The idea proved correct: after checking promising but unsure intersections by means of the Hough transform, a success rate of 100\% was eventually achieved even when disturbance was present, given the last stone put on the goban were fully visible. Even under fair conditions the success rate was close to 100\%, with of without disturbance.
\\This algorithm proved so good mostly because of two reasons:

\SssSec{} Although a disturbance does not look like a stone, it's difficult to tell apart the one from the other if we only stick to empty intersections. Something else is needed in order to decide where the differences lie and help from previously located stones is crucial. For example, let's see what happens employing this algorithm on the picture in figure~\ref{fig:newmethod}:
\begin{figure}[!htb]
  \centering
  \includegraphics[width=\FullWidth]{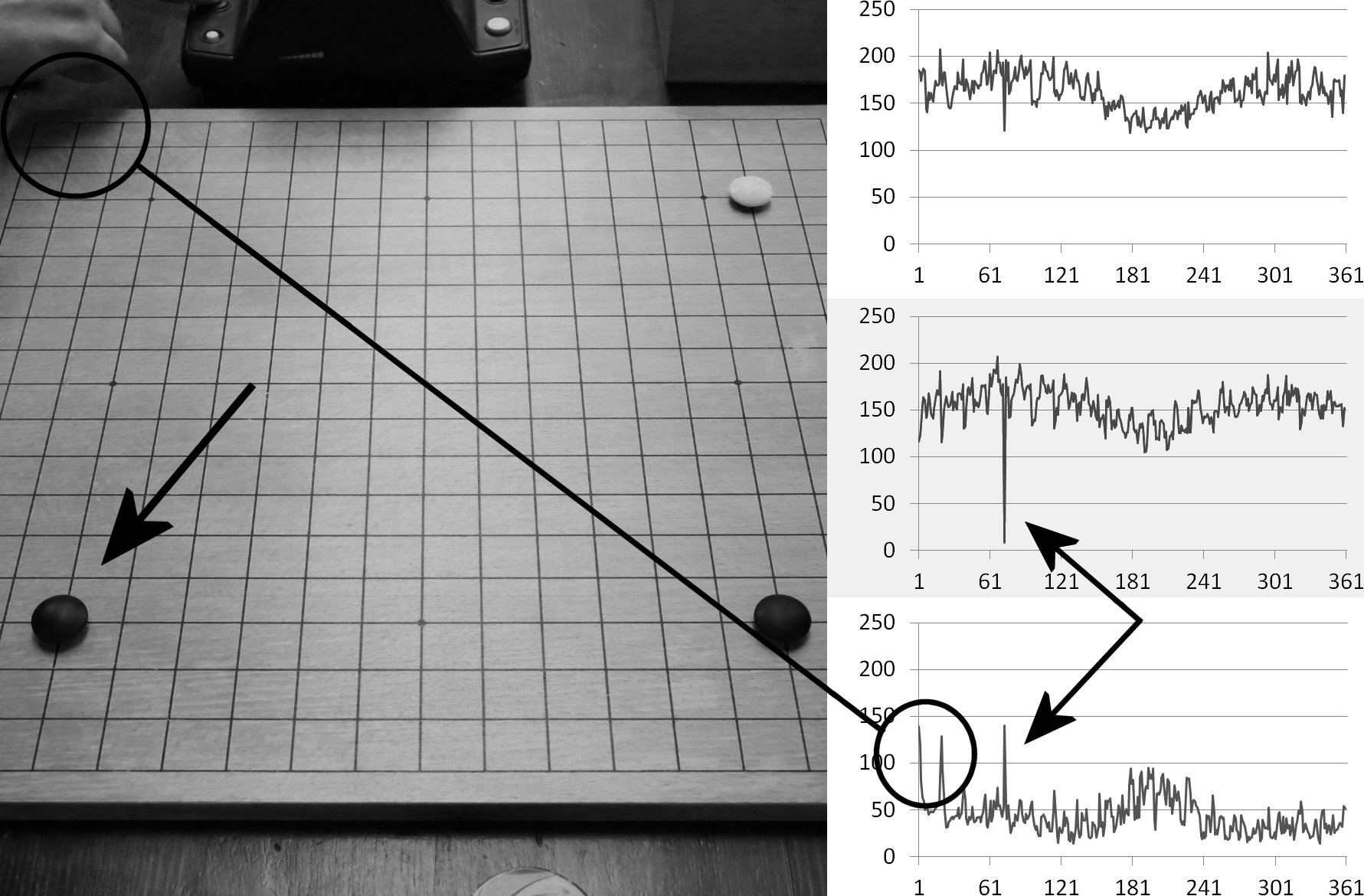}
  \caption{distribution of differences between empty intersections and average white stone (top), average black stone (middle), average empty intersection (bottom).}
  \label{fig:newmethod}
\end{figure}
the distributions on the right show the differences between the local features' mean values for each empty intersection and the corresponding mean values of the ``average'' white stone, the ``average'' black stone, the ``average'' empty intersection respectively. Some intersections don't look ``like empty'' (due to the shadow's disturbance, highlighted by a circle), but only one of them also looks ``like black'', while none at all looks ``like white''. That's more than enough to detect the last stone put on the goban, of course a black one, pointed (together with the corresponding peaks) by an arrow.

\SssSec{} We stated before that the thresholds needed by the ``minimal'' approach make things easier, as isolated peaks can be immediately detected and we need not to determine ``how big'' these peaks should be. But these thresholds are still sensitive to variations in the light or the point of view, disturbance and so on --- hence the limit in the success rate. In the new technique we still need to define ``how similar'' (to those of a stone) or ``how apart'' an intersections' local features must be in order to identify a stone, but these parameters are not affected by light, disturbance and so on, thus making it possible to set them once and for all --- hence the 100\% success rate.

\subsubsection{The details}

\begin{enumerate}
\setlength\itemsep{0em}
\item the intersections are sorted into three groups: empty, white stones, black stones (stone positions, except for the last one put on the goban, are known from previous pictures).
\item for each intersection, the local features values are computed:
\vspace{-2mm}
\begin{itemize}[leftmargin=*]
\setlength\itemsep{0em}
\item around each intersection a circle is scanned, its radius being roughly $1/5$ of the size of a stone, its centre shifted to compensate for projection effects; the area around the real centre, where the grid lines intersect, is let out (see figure~\ref{fig:mezzelune}) as it contains a lot of dark pixels that reduce the intersection's luminance and chrominance, making it dangerously similar to a black stone. The RGB, the hue and the saturation values of all the remaining pixels are then averaged.
\item the intersection's luminance $L$ is then computed by means of the known formula \[L = 0.299 R + 0.587 G + 0.114 B,\]while the chrominance $C$ is the standard deviation of the RGB values. Hue $H$ and saturation are usually part of the pixel attributes and do not require to be computed.
\item starting in the intersection's centre and ending after a complete turn, an upward spiral is swept. Its diameter is about $1/8$ of that of a stone (see figure~\ref{fig:spirals}). For each pixel along the spiral the ``distance'' from the preceding one is computed by means of the formula $\sqrt{(\Delta L)^2 + (\Delta C)^2}$.
\\If the sum of the ``distances'' --- that we call ``disuniformity'' $D$ --- is high, it's likely because around the intersection the grid is fully visible (hence the intersection is probably empty); if it's low, the grid is likely not visible (hence, a stone is probably there).\!\footnote{~A spiral is needed because the shift cannot be applied to a circumference: were it employed, problems would appear on the upper border's intersections where the grid would be missed, thus spawning a low disuniformity whether or not stones were present; were it not, projection effects could not be neglected.}
\begin{figure}[!htb]
  \centering
  \begin{minipage}[t]{\HalfWidth}
    \includegraphics[width=\textwidth]{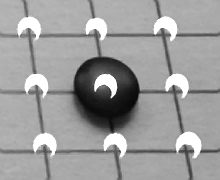}
    \caption{the areas used for computing the intersections' local features.}
    \label{fig:mezzelune}
  \end{minipage}
  \quad
  \begin{minipage}[t]{\HalfWidth}
    \includegraphics[width=\textwidth]{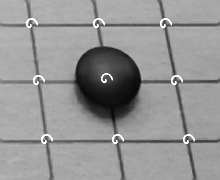}
    \caption{the spirals used for computing the intersections' disuniformity.}
    \label{fig:spirals}
  \end{minipage}
\end{figure}
\end{itemize}
\item for each group, the values of the intersections' local features are averaged. We define these values as the ones pertaining to the ``mean empty intersection'', the ``mean white stone'', the ``mean black stone''.
\item for each \textit{empty} intersection, the differences between the values of its local features and the values of the ``mean empty intersection'', ``mean white stone'' and ``mean black stone'' are computed, then merged together by means of the following formula: \[\sqrt{(\Delta L)^2 + (\Delta C)^2 + (\Delta D)^2 + (\Delta H)^2},\] thus reducing them to only one (at present, saturation is computed but not employed).
\item eventually three discrete functions are built, with domain the set of empty intersections and codomain the differences with the ``mean empty intersection'', the ``mean white stone'', the ``mean black stone'' respectively, computed by means of the formula above (see figure~\ref{fig:newmethod}).
\item the lowest value in the first/second function, ``difference with mean white/black stone'', is checked against three conditions:
\begin{enumerate}
\setlength\itemsep{0em}
\item it does not exceed $2/3$ of the corresponding value of the difference with the ``mean empty intersection'' \textit{(if true, that means the intersection corresponding to the lowest value looks much more like a stone than an empty one)}.
\item it does not exceed $2/3$ of the mean value of the following 15 intersections, assuming the function has been sorted from lowest to highest \textit{(if true, that means this intersection and the ``mean white/black stone'' are much alike)}.
\item its discrete derivative is at least 6 times higher than the mean derivative of the following 15 intersections \textit{(if true, that means it's a peak, hence likely a stone)}.
\end{enumerate}
\item if all three conditions are met, a stone is likely to be found over the intersection corresponding to the function's lowest value. This is double-checked by means of the Hough transform: if a high value is returned, the stone is assumed to be really there (if a low value is returned instead, it is assumed to be a false positive).
\item if only the first condition is met, the intersection is again double-checked by means of the Hough transform. If a very high value is returned, a stone is assumed to be there (thus avoiding a false negative).
\item usually either a white stone or a black stone is found, but both functions are checked nonetheless; if the whole process is repeated on the second to lowest value of both functions other stones could be detected (useful when some pictures are missing or have been discarded).
\end{enumerate}

\section{Conclusion}
\label{sec:Conclusion}

\begin{table*}[!t]
  \setlength\tabcolsep{4pt}
  \centering
  \begin{tabular}{|c||r|r|r|c|c|c|c|c|}
    \hline
    \multirow{3}{*}{\parbox{5ex}{\centering{Game}}} & \multicolumn{4}{c|}{Disturbance} & Stone & \multirow{3}{*}{\parbox{7ex}{\centering{Missing\\pictures}}} & \multirow{3}{*}{\parbox{8ex}{\centering{Duplicate\\pictures}}} & \multirow{3}{*}{\parbox{7ex}{\centering{Moves\\played}}} \\
    \cline{2-5}
    & \multirow{2}{*}{\parbox{12ex}{\centering{none}}} & \parbox{11ex}{\centering{stone fully}} & \parbox{10ex}{\centering{stone~partly}} & stone  & off    & & & \\
    &                                                  & \parbox{11ex}{\centering{visible}}     & \parbox{10ex}{\centering{visible}}      & hidden & centre & & & \\
    \hline
    \hline
    Corsolini-Carta       &           291 &                6 &           1 &  0 & 0 &  1 & 1 & 299 \\
    (friendly game)       & (100.0\%) 291 & (100.0\%)\ \ \ 6 & (100.0\%) 1 &    &   &    &   &     \\
    \hline
    Grazzini-Bevegni      &           189 &               46 &           5 & 12 & 1 &  9 & 5 & 262 \\
    (Florence)            & (100.0\%) 189 &     (100.0\%) 46 &  (60.0\%) 3 &    &   &    &   &     \\
    \hline
    \hline
    Pace-Zingoni          &           239 &               34 &           4 &  0 & 0 &  6 & 0 & 283 \\
    (Florence)            & (100.0\%) 239 &     (100.0\%) 34 &   (0.0\%) 0 &    &   &    &   &     \\
    \hline
    De Lucia-Pace         &           204 &               27 &           3 &  6 & 0 &  6 & 0 & 246 \\
    (Florence)            &  (99.5\%) 203 &     (100.0\%) 27 &  (33.3\%) 1 &    &   &    &   &     \\
    \hline
    Zingoni-Shakhov       &           176 &               19 &           4 &  2 & 0 &  4 & 3 & 205 \\
    (Pisa)                &  (99.4\%) 175 &      (94.7\%) 18 &  (25.0\%) 1 &    &   &    &   &     \\
    \hline
    Pignelli-Albano       &           160 &               54 &           4 &  0 & 0 & 15 & 0 & 233 \\
    (Pisa)                &  (97.5\%) 156 &      (94.4\%) 51 &  (75.0\%) 3 &    &   &    &   &     \\
    \hline
    \hline
    \textit{Imago} test 1 &           224 &               15 &           0 &  0 & 3 &  0 & 2 & 242 \\
                          & (100.0\%) 224 &      (66.7\%) 10 &             &    &   &    &   &     \\
    \hline
  \end{tabular}
  \caption{results. For each game: on the first rows the number of pictures examined, sorted by problem typology; on the second rows the pictures in which the last stone played was correctly detected. Each dataset also includes a photo of the empty goban.}
\end{table*}

We performed some tests to evaluate the success rate achieved by the algorithm under different conditions. After a friendly game in summer 2012, we recorded and analysed some others: first, in December 2012, three games played in the Florence ``Il David'' Tournament, then \textit{Imago}'s first test game\footnote{~The second one was deemed unreliable because too many stones were off-centre (more than 10) and the automatic grid recalibration had to be turned off, as one corner of the grid was too close to the pictures' borders.} and eventually, in March 2015, two games played in the Pisa International Go Tournament, one of which was also filmed in view of the video analysis we hint at later on. In the first table we present the test results; by the way, with only seven test games the exact success rate of the algorithm cannot be determined, and that's why, after a careful scrutiny of the results and the more critical pictures, we estimated an approximate rate presented in a more concise table. Our conclusions were:
\begin{itemize}
\setlength\itemsep{0em}
\item The success rate strictly depends on the goban size (in pixels) and the elevation angle of the point of view. We classified these conditions as shown in the following image:
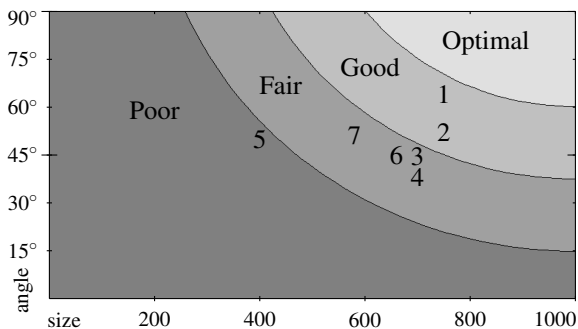
\begin{figure}[!ht]
  \centering
  \setlength{\unitlength}{0.927644\FullWidth}
  \begin{picture}(1.078,0.602)(-0.078,-0.052)
    \put(0,0){\includegraphics[width=0.927644\FullWidth,height=0.510204\FullWidth]{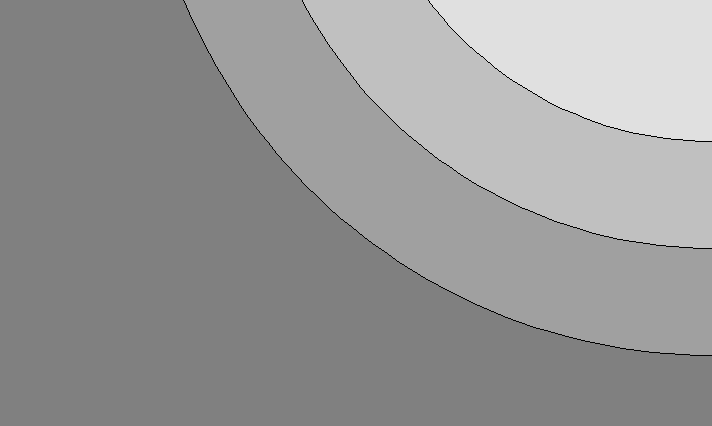}}
    \put(0,0){\line(1,0){1}}
    \put(0,0){\line(0,1){0.55}}
    \multiput(-0.005,0.091667)(0,0.091667){6}{\line(1,0){0.01}}
    \multiput(0.995,0.091667)(0,0.091667){6}{\line(1,0){0.005}}
    \multiput(-0.015,0.275)(0,0.275){2}{\line(1,0){0.03}}
    \multiput(0.985,0.275)(0,0.275){2}{\line(1,0){0.015}}
    \put(-0.05,0.016){\makebox(0,0){\footnotesize{\rotatebox{90}{angle}}}}
    \put(-0.05,0.091667){\makebox(0,0){\footnotesize{$\ang{15}$}}}
    \put(-0.05,0.183333){\makebox(0,0){\footnotesize{$\ang{30}$}}}
    \put(-0.05,0.275){\makebox(0,0){\footnotesize{$\ang{45}$}}}
    \put(-0.05,0.366667){\makebox(0,0){\footnotesize{$\ang{60}$}}}
    \put(-0.05,0.458333){\makebox(0,0){\footnotesize{$\ang{75}$}}}
    \put(-0.05,0.5368){\makebox(0,0){\footnotesize{$\ang{90}$}}}
    \multiput(0.2,-0.005)(0.2,0){5}{\line(0,1){0.01}}
    \multiput(0.2,0.545)(0.2,0){5}{\line(0,1){0.005}}
    \put(0.02825,-0.04){\makebox(0,0){\footnotesize{size}}}
    \put(0.2,-0.04){\makebox(0,0){\footnotesize{$200$}}}
    \put(0.4,-0.04){\makebox(0,0){\footnotesize{$400$}}}
    \put(0.6,-0.04){\makebox(0,0){\footnotesize{$600$}}}
    \put(0.8,-0.04){\makebox(0,0){\footnotesize{$800$}}}
    \put(0.9615,-0.04){\makebox(0,0){\footnotesize{$1000$}}}
    \put(0,0.55){\line(1,0){1}}
    \put(1,0){\line(0,1){0.55}}
    \put(0.75,0.391){\makebox(0,0){$1$}}
    \put(0.75,0.318){\makebox(0,0){$2$}}
    \put(0.70,0.273){\makebox(0,0){$3$}}
    \put(0.70,0.232){\makebox(0,0){$4$}}
    \put(0.40,0.306){\makebox(0,0){$5$}}
    \put(0.66,0.275){\makebox(0,0){$6$}}
    \put(0.58,0.312){\makebox(0,0){$7$}}
    \put(0.83,0.490){\makebox(0,0){Optimal}}
    \put(0.61,0.444){\makebox(0,0){Good}}
    \put(0.44,0.409){\makebox(0,0){Fair}}
    \put(0.20,0.360){\makebox(0,0){Poor}}
  \end{picture}
  \caption{1 Corsolini-Carta, 2 Grazzini-Bevegni, 3 Pace-Zingoni, 4 De Lucia-Pace, 5 \textit{Imago} tests, 6 Zingoni-Shakhov, 7 Pignelli-Albano. Complete datasets are available from \url{http://www.oipaz.net/VideoKifu.html}}
\end{figure}

\item Preconditions: the stones must be where they're supposed to be (stones entirely off-centre cannot be detected) and must be visible.
\item It's difficult to evaluate the success rate when the stones are only partly hidden. Our estimate is based on the test results as well on a careful examination of the few relevant cases.
\item No games were played under optimal conditions, but the success rate can be easily inferred. Poor conditions cannot be evaluated because ``poor'' can mean anything, from ``almost fair'', as in the \textit{Imago} test games that indeed presented a good outcome, to completely useless pictures.
\begin{table}[!htb]
  \setlength\tabcolsep{4pt}
  \centering
  \begin{tabular}{|c|c|c|c|}
    \hline
    \multirow{3}{*}{\parbox{9ex}{\centering{Global\\conditions}}} & \multicolumn{3}{c|}{Disturbance} \\
    \cline{2-4}
    & \multirow{2}{*}{none} & stone fully & \parbox{9.5ex}{stone~partly} \\
    & & visible & visible \\
    \hline
    \hline
    optimal & 100\% & 100\% & 70\%--80\% \\
    \hline
    good & 100\% & 100\% & 60\%--70\% \\
    \hline
    fair & 99\%--100\% & 95\%--100\% & 50\%--60\% \\
    \hline
    poor & \multicolumn{3}{c|}{not evaluated} \\
    \hline
  \end{tabular}
  \caption{success rate per move of the algorithms.}
\end{table}
\end{itemize}

As the above table shows, the algorithm works very well and it's extremely fast, as each picture required about $0.2$ seconds to be processed, plus some necessary pre-processing\footnote{~The required time depends on the image processing library employed: with \href{http://opencv.org/}{OpenCV}, for instance, is almost negligible.} (resizing and converting to B/W in order to make use of the Hough transform).

The test games were analysed by means of a freely distributed program, \textbf{\textit{PhotoKifu}},\!\footnote{~\url{http://www.oipaz.net/VideoKifu.html}} built around the algorithms previously discussed. Yet some aspects of the whole process will be further improved in order to achieve the ultimate goal, that is a real time automatic analysis of a Go game by means of a video stream. For example, the goban tracking algorithm will have to detect larger movements, as we observed that players' fidgeting, combined with tables' swinging, often proves too much for the current procedure; we also noticed that stones are played off-centre more than expected, then exploiting the detection technique's only weakness, which needs to be dealt with. The new program will be called \textit{VideoKifu} and will likely establish a new standard for recording Go games.

\bibliographystyle{alpha}
\bibliography{Carta-Corsolini}
\addcontentsline{toc}{section}{References}

\appendix
\onecolumn
\renewcommand{\thesection}{Appendix}
\section{[Colour images for online publication]}

  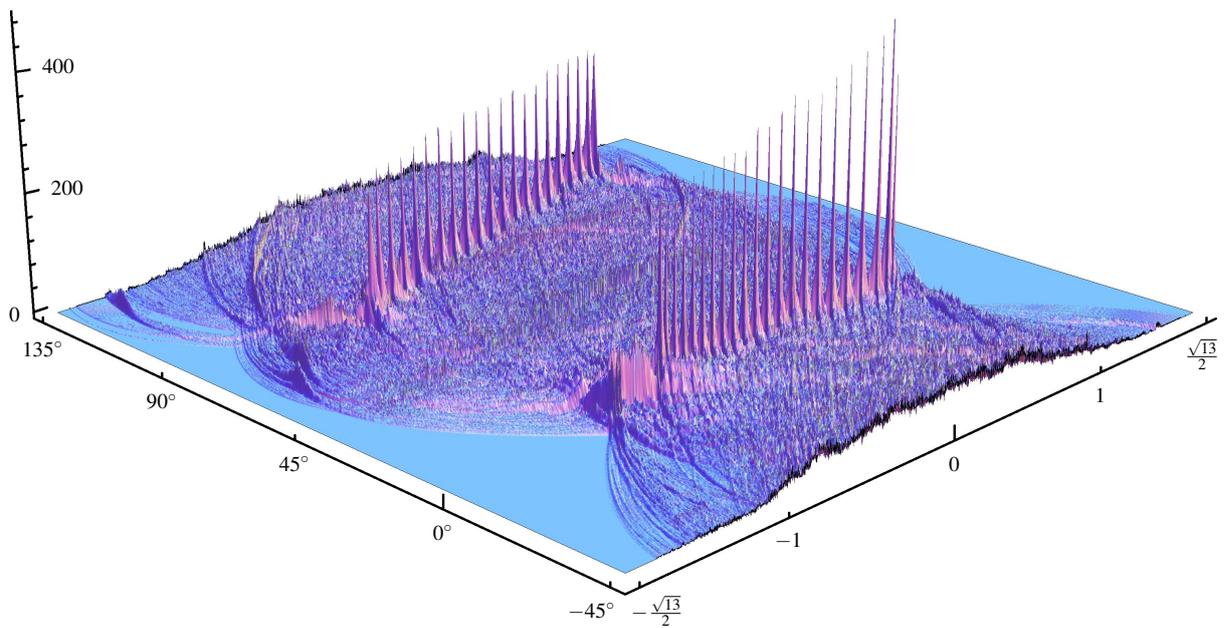
\begin{figure}[!ht]
    \centering
    \setlength{\unitlength}{\FullWidth}
    \begin{picture}(1,0.513215)(0,-0.025)
      \put(0,0){\includegraphics[width=\FullWidth]{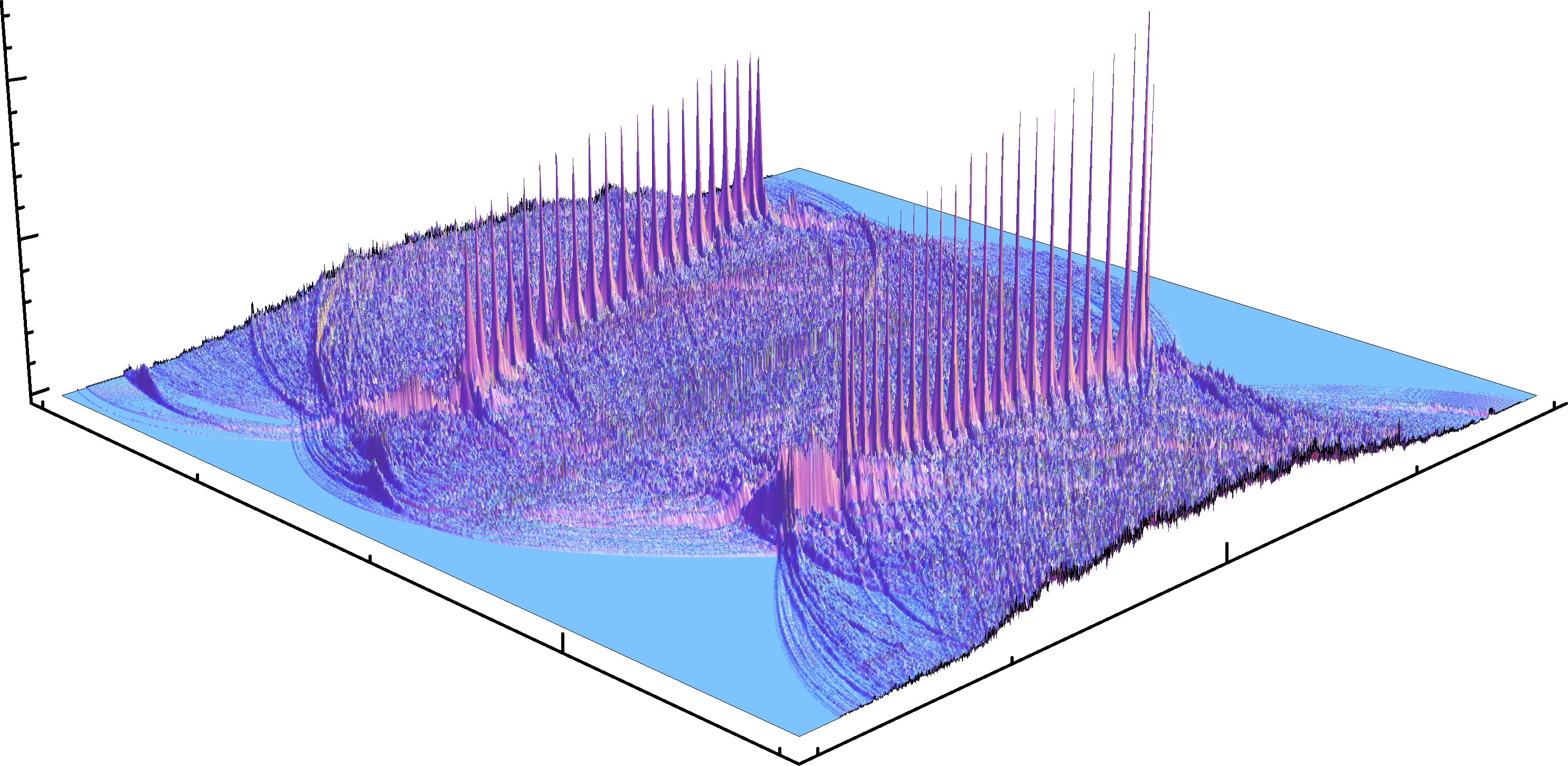}}
      \put(0.039714,0.441956){\makebox(0,0){\footnotesize{$400$}}}
      \put(0.047121,0.340272){\makebox(0,0){\footnotesize{$200$}}}
      \put(0.003529,0.234452){\makebox(0,0){\footnotesize{$0$}}}
      \put(0.026936,0.205795){\makebox(0,0){\footnotesize{$\ang{135}$}}}
      \put(0.125589,0.161818){\makebox(0,0){\footnotesize{$\ang{90}$}}}
      \put(0.235690,0.109966){\makebox(0,0){\footnotesize{$\ang{45}$}}}
      \put(0.358586,0.052391){\makebox(0,0){\footnotesize{$\ang{0}$}}}
      \put(0.481970,-0.012929){\makebox(0,0){\footnotesize{$\ang{-45}$}}}
      \put(0.536548,-0.012929){\makebox(0,0){\footnotesize{$-\frac{\sqrt{13}}{2}$}}}
      \put(0.645118,0.045657){\makebox(0,0){\footnotesize{$-1$}}}
      \put(0.782492,0.109630){\makebox(0,0){\footnotesize{$0$}}}
      \put(0.903367,0.166869){\makebox(0,0){\footnotesize{$1$}}}
      \put(0.987745,0.201462){\makebox(0,0){\footnotesize{$\frac{\sqrt{13}}{2}$}}}
    \end{picture}
    \caption{enlarged colour version of figure~\ref{fig:HT_HoughSpace}.}
  \end{figure}

\begin{figure}[!ht]
  \centering
  \includegraphics[width=\FullWidth]{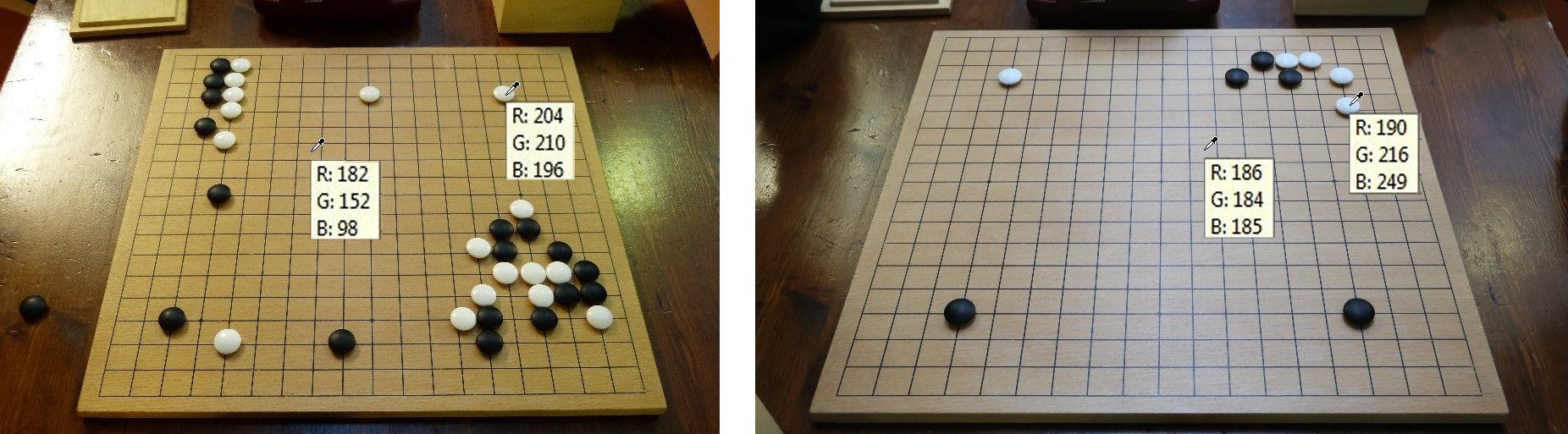}
  \caption{an example of what is discussed in paragraph~\hyperref[par:Features]{\ref*{sssec:Criteria}.\ref*{par:Features}}. Same goban, different light: on the left, natural light (the surface looks colourful), on the right, artificial (not only the surface looks grey, but the white stones look more colourful).}
\end{figure}

\begin{figure}[!ht]
  \centering
  \includegraphics[width=\FullWidth]{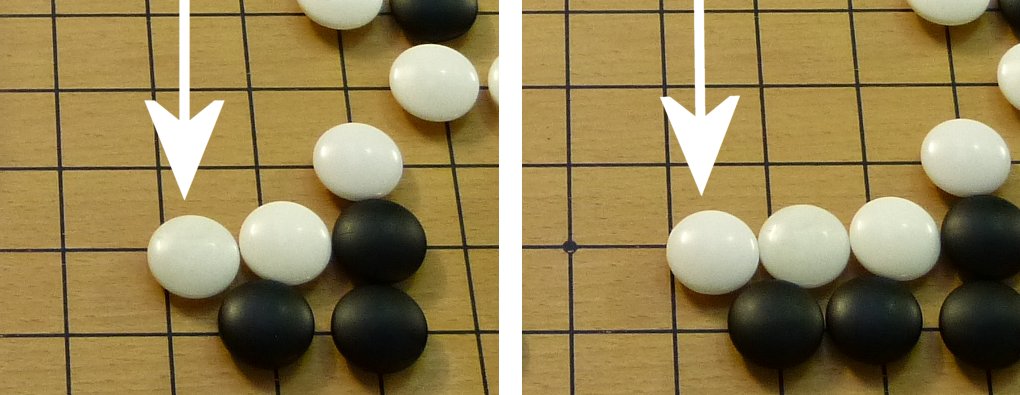}
  \caption{on the left a barely detectable off-centre stone, on the right a stone so off-centre it cannot be detected (see section~\ref{sec:Conclusion}).}
\end{figure}

\end{document}